\documentclass[lettersize,journal]{IEEEtran}
\usepackage{cite}
\usepackage{amsmath,amssymb,amsfonts}
\usepackage{graphicx}
\usepackage{textcomp}
\usepackage{xcolor}
\usepackage{wrapfig}
\usepackage{url}   
\usepackage{algorithmic}
\usepackage{algorithm}
\usepackage{multirow}
\usepackage{xspace}
\usepackage{booktabs}       
\usepackage{amsfonts}       
\usepackage{nicefrac}       
\usepackage{microtype} 
\usepackage{tikz}
\usepackage[capitalize,noabbrev]{cleveref}
\usepackage{subcaption}
\usepackage{makecell}
\usepackage[table]{xcolor}
\def\BibTeX{{\rm B\kern-.05em{\sc i\kern-.025em b}\kern-.08em
    T\kern-.1667em\lower.7ex\hbox{E}\kern-.125emX}}
\begin{document}

\title{Geometric Flow enhanced Graph Coarsening
}

\author{
        Chaoqun~Fei,
        Guoxuan~Li,
        Tinglve~Zhou,
        Chuanqing~Wang,
        Yangyang~Li\\
        \thanks{C.Q. Fei and T.L. Zhou are with the School of Artificial Intelligence, South China Normal University, Foshan 528225, China (e-mail: cqfei@m.scnu.edu.cn; tingluezhou@m.scnu.edu.cn)}
        \thanks{G.X. Li, C.Q W and Y.Y. Li is with the State Key Laboratory of Mathematical Sciences, Academy of Mathematics and Systems Science, Chinese Academy of Sciences, Beijing 100190, China (e-mail: yyli@amss.ac.cn)
        \emph{ (Corresponding author: Yangyang Li.)}}
}

\markboth{Journal of \LaTeX\ Class Files,~Vol.~14, No.~8, August~2021}%
{Shell \MakeLowercase{\textit{et al.}}: A Sample Article Using IEEEtran.cls for IEEE Journals}

\maketitle

\begin{abstract}
Recently, researchers have proposed a graph pooling operation, akin to the pooling process in conventional convolutional neural networks (CNN), aimed at reducing the computation cost of Graph convolutional neural networks (GCNNs). While most GCNN-based methods treat graph pooling as a node clustering problem and propose learning a cluster assignment matrix, existing clustering-based pooling methods tend to focus solely on the rough topology information of graphs, neglecting the exploitation of higher-order mutual connections among neighbors. In terms of message passing on graph, the ease of information passing on edges reflects the closeness between neighboring nodes, which significantly relies on the interconnectivity among neighbors. 
In this study, we address this gap by considering such local connection information and introducing a novel graph pooling method named RicciPool. We introduce discrete graph curvature, particularly Ollivier-Ricci curvature, as a measure of higher-order connectivity around an edge. Subsequently, we construct an Ollivier-Ricci flow formula to reweigh edge weights, leveraging the crucial information provided by Ricci curvature, particularly vital for extracting clusters in graphs. Building upon this foundation, we utilize the spectral clustering technique to learn a new cluster assignment matrix. Experimental results on multiple bioinformatics protein datasets and social networks underscore the effectiveness of our proposed method. 
\end{abstract}


\section{Introduction}
\label{Introduction}

Over the past decade, Convolutional Neural Networks (CNNs) have revolutionized fields such as computer vision and other domains characterized by Euclidean data structures with uniform, grid-like arrangements. Despite their significant success, fully exploiting the geometric or topological structures of non-Euclidean domains remains challenging. Non-Euclidean domains include data structures represented by sub-manifolds or graphs~\cite{Monti2017}, encompassing areas like 3D graphics, social networks, biological systems, and web traffic. Traditional convolutional operations in CNNs rely on the Euclidean metric to compute linear weighted sums, but this metric is limited in capturing the nonlinear feature distribution of non-Euclidean structured data~\cite{bronstein2017geometric}.

Graph convolutional neural networks  (GCNNs)~\cite{Joan2013}\cite{defferrard2016}\cite{Xu2019}\cite{Henaff2015}\cite{Wu_and_Pan_2021} are a novel deep learning method that extends Convolutional Neural Networks (CNNs) from traditional Euclidean domains to graph-structured data. GCNNs leverage the topology of graphs to learn node and graph embeddings by propagating information between nodes and their neighbors. Unlike traditional CNNs, GCNNs can handle non-uniform and irregular data structures such as social networks, molecular structures, and knowledge graphs. In recent years, GCNNs have demonstrated strong performance in tasks such as node classification, link prediction, and graph classification, becoming a key tool for handling graph data~\cite{cai2018}. Node classification~\cite{gao2019b}\cite{Petar2018} and link prediction~\cite{cai2020}\cite{zhang2018linkprediction} focus on learning representations at the node level using features and topology, while graph classification~\cite{zhang2018graphclassification} targets graph-level representation learning to predict labels for entire graphs. These methods have demonstrated promising results in various graph-related tasks. Despite their promising results in various tasks, traditional GCNNs face challenges in efficiently computing large-scale graphs, characterized by many nodes or links~\cite{will2017}. In conventional CNNs, pooling operations effectively reduce feature dimensionality and computational costs for large-scale grid-structured data, allowing for deeper network designs and mitigating overfitting. However, these pooling techniques cannot be directly applied to graph-structured data due to its non-Euclidean nature. Consequently, several graph-based pooling methods have been developed in conjunction with GCNNs~\cite{selfpool}\cite{diffpool}\cite{structpool}\cite{Grattarola_and_Zambon}\cite{Wang_and_Chang_2022}\cite{Bianchi_and_Grattarola_2022}.

\begin{figure}
	\centering
	\vspace{-0mm}
	\includegraphics[width=1.0\linewidth]{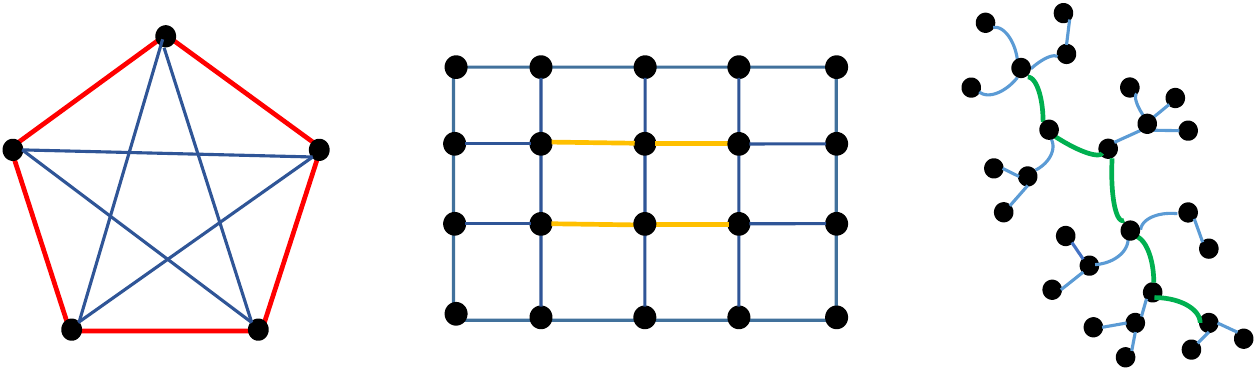}
	\vspace{-4mm}
	\caption{Example of Ricci curvature on graphs. From left to right, the Ricci curvature of the red edges are positive, the orange edges are zero, and negative for the green edges in tree-structure graph, respectively. However, all edges have same weight $1$.}
	\label{fig:graph_curvature}
	\vspace{-1mm}
\end{figure}

Mainstream methods treat graph pooling as a node clustering task, where each cluster corresponds to a supernode in the pooled graph, as demonstrated in DiffPool~\cite{diffpool}, MinCutPool~\cite{spectralpool}, and StructPool~\cite{structpool}. These methods focus on learning a new cluster assignment matrix. Topologically, nodes with shorter path distances in a graph should have a higher probability of being assigned to the same cluster. Accurately capturing pairwise node similarities is crucial for effective graph pooling. However, current methods primarily focus on node-specific attributes or coarse topological information, such as node degree, which fails to accurately reflect spatial similarities in node distribution. For instance, different graph topologies can have same node degree distribution, as shown in Fig.~\ref{fig:graph_curvature}. 

In terms of message passing on a graph, the ease of information transfer on edges reflects the closeness between neighboring nodes. Adjacency nodes in a well-connected community have highly overlapped neighborhood sets and numerous shortcuts, facilitating easier message passing compared to a hierarchically structured community. However, existing graph pooling methods do not distinguish between these node connection structures, often setting edge weights equally and overlooking local similarities. To address these differences, we draw inspiration from recent studies on graph curvature~\cite{Lin2011} to explore a new metric for reweighting graph edges. This metric leverages curvature information to ensure that easier message passing on an edge corresponds to shorter edge distances.

Discrete graph curvature measures the deviation of a graph's neighborhood domains from a "flat" domain, such as a grid graph. Two prominent discrete curvature measures, Ollivier-Ricci curvature~\cite{ollivier2009ricci}\cite{ollivier2010} and Forman curvature~\cite{Forman2003}, have recently garnered attention. Ollivier-Ricci curvature, based on optimal transportation theory, has been applied to various structural tasks~\cite{sia2019}\cite{ni2018network}, whereas Forman curvature, derived from the graph Laplacian, is computationally simpler and faster but lacks geometric interpretability. In this work, we utilize Ollivier-Ricci curvature to investigate the local geometric structure of node distribution, measuring the difficulty of message passing on edges. Essentially, Ollivier-Ricci curvature assesses the connection density of subgraphs, which can be categorized into positive, zero, and negative values. Positive curvature signifies a dense, clique-like graph, zero curvature represents a grid-like graph, and negative curvature indicates a sparse, tree-like graph~\cite{li2022curvature}, as shown in Fig.~\ref{fig:graph_curvature}.

In this paper, we introduce a novel method called RicciPool, a \textsl{Graph Pool via Ricci Flow}. We use the edge weight matrix as a distance metric to quantify message passing distances between nodes. Initially, like other GCNN-based methods, the edge weight matrix is an unweighted adjacency matrix with all edge weights set to one. We develop an Ollivier-Ricci flow formula on the graph, leveraging the corresponding Ollivier-Ricci curvature~\cite{ni2019community} to iteratively adjust the edge weight matrix for better discrimination. This flow guides graph curvature evolution through a nonlinear partial differential equation, expanding negative curvature edges and contracting positive curvature edges. This iterative process condenses nodes connected by overlapping edges and stretches tree-structured edges. Following the graph Ricci flow evolution, we adopt the spectral clustering technique~\cite{Yu2003}, similar to MinCutPool~\cite{spectralpool}, combined with GCNN to derive a new cluster assignment matrix. Consequently, nodes in clique-like subgraphs are more likely to be clustered together. During pooling, nodes within each cluster are aggregated into a new supernode, and the edge weight matrix and associated node features are updated using the learned assignment matrix. Theoretically, this preserves the original graph's topology. After pooling, the attribute features of supernodes incorporate local topology structures and the original node set's attributes.

The contributions of this paper are summarized as follows:
\begin{enumerate}
    \item Enhance Ollivier-Ricci flow for graph pooling via a curvature-driven coarsening strategy that dynamically adjusts edge weights and preserves structural bottlenecks by focusing on high positive curvature connections.
	\item Propose a novel spectral clustering-based topology-preserving graph pooling method. By learning a structure-aware node assignment matrix, this approach systematically preserves the global topological features of the original graph during coarsening.
    \item Extensive experiments on benchmark graph classification datasets demonstrate the effectiveness and competitiveness of RicciPool.

\end{enumerate}

The remainder of this paper is organized as follows. In Section~\ref{related},  we provide an overview of related works on graph pooling in Graph Neural Networks and discrete graph curvature.  Section~\ref{sec:preli} presents the preliminaries, including basic definitions and theories. The proposed graph pooling method based on Ricci flow is detailed in Section~\ref{sec:method}. Section~\ref{sec:exp} presents the experiments and evaluation results. Finally, we conclude our work in Section~\ref{sec:conc}.

\section{Related Works}\label{related}

\subsection{Graph Pooling in GNNs}

In recent years, there has been significant development in GNN-based graph pooling methods, which can be primarily categorized into two types based on distinct pooling principles.
The first type involves methods based on node selection principles. These methods establish a scoring metric to quantify node importance and subsequently rank nodes according to these scores. Examples include SORTPool~\cite{zhang2018graphclassification}, TOPKPool~\cite{gao2019}, and SAGPool~\cite{selfpool}. However, their exclusive consideration of the global topological structure of a graph neglects potentially significant local connections contained within ignored nodes, which are lost during pooling.

The second type, termed graph coarsening, treats graph pooling as a node clustering problem based on graph topology. DIFFPool~\cite{diffpool} constructs a multi-level GNN network with pooling layers in each hidden layer. StructPool~\cite{structpool} utilizes conditional random fields to learn a cluster assignment matrix, while MinCutPool~\cite{spectralpool} frames the node clustering task as a minCUT problem. Additional approaches include EigenPool~\cite{eigenpool} and HaarPool~\cite{haarpool}, which consider both topology information between nodes and the embedding of the entire graph. However, few of them explicitly differentiate between different pairwise connected structures of nodes.

\subsection{Discrete Graph Curvature}

Recent attention has been drawn to various proposals of discrete curvature on weighted or unweighted graphs, notably Ollivier-Ricci curvature and Forman curvature. Ollivier-Ricci curvature, rooted in optimal transportation theory, exhibits a more geometric nature. It has been integrated into graph neural networks to recalibrate different message channels~\cite{li2022curvature}\cite{ye2019curvature} and applied in network alignment~\cite{ni2018network} and community detection~\cite{sia2019}\cite{ni2019community}. In contrast, Forman curvature, grounded in the graph Laplacian, offers faster computation and has been used to analyze large-scale graphs such as complex networks~\cite{weber2018}\cite{sreejith2016}\cite{samal2018} and bioinformatics-related graphs~\cite{sandhu2015}. Forman curvature flow has also been proposed for community detection~\cite{weber2016}\cite{weber2017}. However, comparative studies have favored Ollivier-Ricci curvature for graph structure detection~\cite{ni2019community}. Thus, in this study, we choose to utilize Ollivier-Ricci curvature for reweighing edge weights.

\section{Preliminaries}\label{sec:preli}   

In this section, we aim to elucidate the fundamental concepts pertinent to our approach. Firstly, we delineate the foundational framework of graph convolutional neural networks. Subsequently, we expound upon the pooling operation within the realm of graph-structured domains. Lastly, we provide the formal definition of Ricci flow on a continuous manifold. Formally, let's consider a graph $G = \{V, E\}$ with an unweighted adjacent matrix $A \in \left(0,1\right)^{N\times N}$ and a feature matrix $X \in \mathbb{R}^{N\times F}$ pertaining to the node set $V$. Here $E$ denotes the edge set, $F$ signifies the feature dimensions, and $N$ represents the total number of nodes within the graph.

\subsection{Graph Convolutional Neural Networks}
\label{GCNN}
A graph can be characterized by its adjacency matrix and node feature matrix. Graph Convolutional Neural Networks (GCNNs) leverage these matrices to learn feature embeddings at either the node or graph level, employing both spectral and spatial methods. Spectral methods~\cite{defferrard2016}\cite{Henaff2015} utilize the graph Laplacian operator to devise convolution operations. However, the convolution filters learned through spectral methods are tailored to specific graphs and thus lack universality to address diverse graph structures. In contrast, spatial approaches~\cite{Xu2019}\cite{Petar2018} in GCNNs typically aggregate neighbor representations to formulate convolution operations for updating the features of individual nodes~\cite{zhang2018graphclassification}, formally defined as follows:
\begin{align}\label{eq:2}
    X^{i+1} = f\left(D^{-1}\hat{A}X^i T^i\right),
\end{align}
where $\hat{A}=A+I$, $D_{j,j}=\sum_{k}\hat{A}_{jk}$, $X^i$ is the corresponding node feature matrix after $i$-th convolution operation, $T^i$ is a trainable weight matrix to perform feature transformation, and $f\left(\cdot\right)$ represents a non-linear activation function. In our model, we use this version of GCNNs in Equation~\eqref{eq:2} for the basic convolution operation.

\subsection{Graph Pooling}

Similar to the pooling operation in CNNs, graph pooling aims to reduce the size of nodes and edges within a graph. Formally, a graph pooling operator should produce a new feature matrix $X^{pool}\in \mathbb{R}^{K\times F'}$ and an adjacency matrix $A^{pool}\in \mathbb{R}^{K\times K}$ in the pooling layer, typically with $N\ge K$~\cite{spectralpool}. To achieve this objective, the graph pooling operation endeavors to learn a bilinear transformation matrix $P\in \mathbb{R}^{N\times K}$ between two adjacent convolutional layers, satisfying:
\begin{align}\label{eq:3}
    A^{pool} = P^TAP;\quad X^{pool}=P^TX.
\end{align}
The new adjacent matrix $A^{pool}$ should inherit and preserve the global topological structure of the initial $A$. In clustering-based graph pooling methods, the transformation matrix $P$ is referred to as the cluster assignment matrix.

\subsection{Ricci Flow}\label{subsec:ricci}

In 1982, Richard Hamilton first proposed Ricci flow equation~\cite{hamilton1982three}, a parabolic partial differential equation deforming the Riemannian metric by its Ricci curvature:
\begin{align}
    \frac{\partial}{\partial t} g_{ij} = -2R_{ij}.
\end{align}
Here, $g_{ij}$ denotes the Riemannian metric defined on a continuous manifold, and $R_{ij}$ represents the corresponding Riemannian curvature. The Ricci flow can be conceptualized as a heat equation for the Riemannian metric. Through Ricci flow, the volumes of neighborhoods with positive sectional curvature generally contract, whereas those with negative sectional curvature expand.  
In Riemannian geometry, regions with significant positive curvature are generally more densely packed than those with negative curvature~\cite{ni2019community}.

Inspired by Hamilton's Ricci flow, we propose a geometric evolution equation on graphs, termed the Ollivier-Ricci flow, for detecting clique-like subgraphs and tree-like subgraphs. The Ollivier-Ricci flow interprets the edge weight matrix $w$ as a discrete distance metric defined on a graph, analogous to the Riemannian metric $g_{ij}$ in continuous manifold and evolves the edge weights over time: the edge weights of densely connected structures (positive Ricci curvature) are diminished, while those of sparsely connected structures (negative Ricci curvature) are enlarged~\cite{ni2019community}.

\section{Proposed Method}\label{sec:method} 

Firstly, we propose leveraging the curvature information of a graph to construct an Ollivier-Ricci flow on the graph, facilitating the iterative update of the edge weight matrix. Subsequently, based on the updated weight matrix, we partition the node set into $K$ disjoint clusters. 
Finally, we utilize both GCNN and a learned cluster assignment matrix to condense the graph into a more compact version within the graph pooling layer.

\subsection{Ollivier-Ricci Curvature}\label{subsec:curvature}

Similar to the Ricci curvature in differential manifold, Ollivier-Ricci curvature as its discrete counterpart on graphs was initially introduced by Y. Ollivier utilizing the optimal transportation theory. Ollivier defines the curvature of an edge by assigning a probability measure to each node at both ends of the edge, with the Ollivier-Ricci curvature of the edge determined by the optimal transportation cost between these two probability measures.

Given a graph $G = \left(V, E, A\right)$, we define an edge distance metric on $G$, termed the edge weight matrix $w$. Here, we consider the unweighted adjacency matrix $A\in (0,1)^{N\times N}$ as the initial edge weight matrix. For each node $x\in V$, we denotes a probability distribution $m_x$ on the neighbor node set of $x$~\cite{Anca2009} (i.e., the nodes connected to $x$). The discrete Ricci curvature $R_{xy}$ on edge $xy\in E$ is formally represented as~\cite{ollivier2009ricci}:
\begin{align}
    R_{xy} = 1-\frac{W\left(m_x, m_y\right)}{d\left(x, y\right)}.
    \label{eq:curvature}
\end{align}
The Wasserstein distance $W\left(m_x, m_y\right)$  here is defined as the optimal transportation cost required to move from $m_x$ to $m_y$. The term $d\left(x,y\right)$ signifies the shortest path between nodes $x$ and $y$, as outlined in Section~\ref{subsec:flow}. The measure $m_x$ is represented as~\cite{ni2019community}:
\begin{equation}
    m_x^\alpha (x_i)= \left\{ 
    \begin{aligned}
    &\alpha & \mbox{if}\quad x_i =x\\
    &(1-\alpha)/|\pi(x)| &  \mbox{if}\quad x_i \thicksim x \\
    & 0 & \mbox{otherwise},
    \end{aligned}
    \right.
\end{equation}
where $\alpha$ is a parameter within $[0,1]$, $\pi(x)$ denotes the neighbor set of $x$ and $|\pi(x)|$ denotes nodes size in $\pi(x)$. In this context, we set $\alpha = 0.5$ following to~\cite{ni2019community}.

\subsection{Ollivier-Ricci Flow}\label{subsec:flow}

Ollivier-Ricci flow defined on a graph can be seen as a discrete version of the Ricci flow on a continuous manifold. It regularize the edge weight matrix $w$ on a graph by evolving the Ollivier-Ricci curvature $R$. The equation of Ollivier-Ricci flow is shown below:
\begin{align}\label{eq:7}
    \frac{d}{dt}d^t\left(x,y\right) = -R^t_{xy}d^t\left(x,y\right).
\end{align}
where, $d(x,y)$ denotes the shortest path distance between nodes $x$ and $y$.

Here, we define the concept of the shortest path $d$ in a graph. Let $G=\left(V,E,w\right)$ be a weighted graph where $w_{xy}>0$ for every edge. The shortest path $d$ between nodes in the set $V$ is determined as follows:
\begin{align}
    d\left(v,v'\right)=
    \underset{w}{\min}\left\{\sum_{i=1}^{n}w_{s_i,s_{i+1}}: s_i \in V, s_0 =v, s_{n+1}=v'\right\}.
\end{align}
The shortest path is calculated over all edge paths connecting $v$ to $v'$. We refer to $d$ as the distance metric induced by the edge weight $w$. By definition, for any three vertices $v, v'$, and $v'' \in V$, it holds that $d(v, v') + d(v', v'') \ge d(v, v'')$, adhering to the metric rule.

For the initial unweighted edge weight matrix, denoted as $w^0 = A \in (0,1)^{N \times N}$, the shortest path $d(x,y)$ between any two neighboring nodes is equal to the edge weight $w^0(x,y)$ and set to one.

According to the Ollivier-Ricci flow described in Eq.~\eqref{eq:7}, we restrict our attention to the edge curvature between neighboring nodes. For such nodes, the shortest path distance $d(x,y)$ coincides with the edge weight $w_{xy}=d(x,y)$. In each iteration, both the edge weight matrix $w$ and the Ollivier-Ricci curvature are updated concurrently via the following flow process:
\begin{align}
    w_{xy}^{t+1}& = d^t\left(x,y\right)\left(1-\eta \cdot R_{xy}^t\right);\label{eq:8}\\
    R_{xy}^{t+1}& =1-\frac{W^{t+1}\left(m_x, m_y\right)}{d^{t+1}\left(x,y\right)}.\label{eq:81}
\end{align}
At the $(t+1)$-th iteration, $w_{xy}^{t+1}$ represents the weight of edge $xy$, $R_{xy}^{t+1}$ denotes the Ollivier-Ricci curvature of edge $xy$, and $d^t\left(x,y\right)$ is the corresponding shortest path distance between $x$ and $y$ induced by the edge weight matrix $w^t$. Initially, $w^0_{xy} = A_{xy}$ and $d^0\left(x,y\right)=w^0_{xy}$.

Note that according to Eq.~\eqref{eq:8}, this discrete Ricci flow process enlarges the edge weights with negative edge curvature while shrinking the weights of the positive curvature edges, if we consider the weight of an edge as the edge distance. Eventually, nodes connected by the clique-like structural edges are condensed and the tree-like structural edges are stretched. This effect allows for the easy separation of closely connected subgraphs through a simple strategy, such as removing edges with weights greater than a specified threshold. The detailed algorithm framework of the Ollivier-Ricci flow process is illustrated in Algorithm~\ref{alg:example}. In Step 3 of the algorithm, we define the stopping criterion for the Ricci flow iteration: the process terminates when the Ollivier-Ricci curvature of all graph edges stabilizes (i.e., ceases to change or changes by less than a predefined threshold $\delta$) between consecutive iterations. The core function of this step is to verify satisfaction of this criterion. Upon meeting the condition, the Ricci flow iteration halts immediately.

\begin{algorithm}[tb]
    \caption{Ollivier-Ricci Flow}
  \label{alg:example}
 \begin{algorithmic}
   \STATE {\bfseries Input:} the initial adjacency matrix $A\in(0,1)^{N\times N}$, \\
  \quad \quad \quad a parameter $\eta$ and a threshold $\delta$.\\
   \STATE 1. Initialize that $w^0 = A$, $d^0(x,y) = w^0_{xy}$. 
   \STATE 2. Calculate the Ollivier-Ricci curvature of each edge using Eq.~\eqref{eq:curvature}.
   \STATE 3. {\bfseries while} $|R_{xy}^{t+1}-R_{xy}^t|>\delta$ {\bfseries do}
   \STATE 4. \quad Update the edge weight by Eq.~\eqref{eq:8};
   \STATE 5. \quad Update the Ollivier-Ricci curvature by Eq.~\eqref{eq:81}.
   \STATE 6. {\bfseries end while}
   \STATE {\bfseries Output:} the updated edge weight matrix $w^R$.
 \end{algorithmic}
\end{algorithm}

\begin{figure*}
	\centering
	\includegraphics[width=1.0\linewidth]{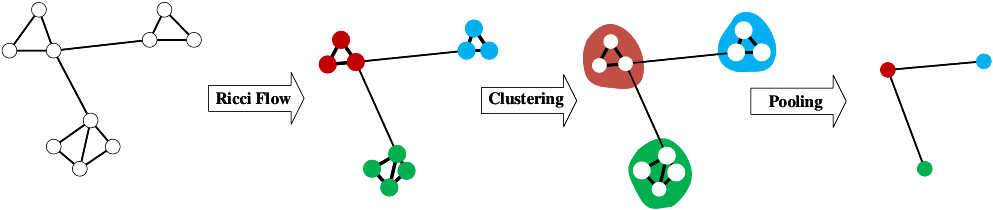}
	\caption{Ollivier Ricci flow based graph pooling model.}
	\label{fig:framework}
\end{figure*}

\subsection{Nodes Clustering}\label{subsec:cluster}
 Building upon the minCUT problem in spectral clustering~\cite{Yu2003}, we employ spectral techniques to learn a node clustering assignment matrix for graph partitioning. Specifically, given a graph $G=(V,E,A)$, with $N$ nodes and a predefined number of clusters $K$ ($K<N$). $V$ is the node set of $G$. The assignment principle in our approach is to minimize the total edge weights within clusters while maximizing the number of intra-cluster edges, and simultaneously minimize the number of inter-cluster edges while maximizing inter-cluster edge weights. This leads to the following loss function (Eq.\eqref{eq:9}):
\begin{align}\label{eq:9}
    \mathcal{L} = \frac{1}{K}\sum_{k=1}^{K}\left\{\frac{weights\left(V_k\right)}{weights \left(V/V_k\right)}+\frac{degree \left(V/V_k\right)}{links\left(V_k\right)}\right\}.
\end{align}

For the $k$-th cluster $V_k$, we define the following metrics:
\begin{itemize}
	\item $V/V_k$: The set of all nodes not in cluster $V_k$.
	\item $weights(V_k)$: The sum of all edge weights within $V_k$.
	\item $links(V_k)$: The number of edges within $V_k$.
    \item $degree(V/V_k)$: The number of edges connecting $V_k$ to the rest of the graph $(V/V_k)$.
    \item $weights(V/V_k)$: The total weight of all edges between $V_k$ and $V/V_k$.
\end{itemize}

To incorporate geometric information from graph curvature flow, we introduce the Ollivier-Ricci flow evolved weight matrix $w^R$, obtained by iteratively updating the initial weight matrix $A$. Let $P\in \{0,1\}^{N\times K}$ be the cluster assignment matrix, where $P_{ij}=1$ if node $i$ belongs to cluster $j$, and $0$ otherwise, with $P_k$ denoting its $k$-th column. The degree matrices are defined as $D = diag\left(A\textbf{1}_N\right)$ and $D^R=diag(w^R\textbf{1}_N)$ for the original and evolved graphs, respectively. The cluster-wise objective is formulated as Eq.\eqref{eq:10}:
\begin{align}\label{eq:10}
    \underset{P}{\arg\min} \frac{1}{K}\sum_{k=1}^{K}\left\{\frac{P_k^T w^R P_k}{P^T_kD^R P_k} + \frac{P_k^T D P_k}{P_k^T A P_k}\right\}.
\end{align}
For the $k$-th cluster $V_k$, we define the following equivalent matrix forms:
\begin{itemize}
    \item First Term ($P_k^Tw^RP_k$): The equivalence between $P_k^Tw^RP_k$ and $weights(V_k)$ is exact. The quadratic form $P_k^Tw^RP_k$ computes the sum of edge weights for all pairs of nodes in cluster $k$. In an undirected graph without self-loops, this equals $2\times weights(V_k)$, as each internal edge is counted twice. Minimizing $P_k^Tw^RP_k$ is thus functionally identical to minimizing weights ($V_k$) for the purpose of optimization.
    \item Second Term ($P_kD^RP_k$): The denominator term $P_kD^RP_k$ is a surrogate for $weights(V/V_k)$.
    \begin{align*}
    P_k^T D^R P_k &= \sum_{i \in V_k} (\sum_{j \in V_k} w^R_{ij} + \sum_{j \in V/V_k} w^R_{ij})\\
    &= \sum_{i \in V_k} \sum_{j \in V_k} w^R_{ij} + \sum_{i \in V_k} \sum_{j \in V/V_k} w^R_{ij}\\
    &=(2\times weights(V_k)) + weights(V/V_k).
    \end{align*}
    Consequently, maximizing $P_kD^RP_k$ is functionally equivalent to maximizing $weights(V/V_k)$.
    \item The metrics $link(V_k)$ and $degree(V/V_k)$ represent the unweighted counterparts of $weights(V_k)$ and $weights(V/V_k)$, respectively, focusing on edge counts rather than the sum of their weights.
\end{itemize}

Using the matrix identity for any matrix $X$:
\begin{equation*}
    \sum_{k=1}^{K}P_k^T XP_k = tr(P^TXP)
\end{equation*}
where the equality follows from the observation that $(P^TXP)_{kk}=P_k^TXP_k$ and the trace sums the diagonal entries, i.e., $tr(P^TXP)=\sum_{k}^{K}(P^TXP)_{kk}$. This allows us to transform the component-wise sums into global trace terms. Denote these common values as $\Delta_{min} = \min \{P_k^T D^R P_k, P_k^T A P_k\}$ and $\Delta_{max} = \max \{P_k^T D^R P_k, P_k^T A P_k\}$ for all $k$. Then, we derive the following inequalities:
\begin{equation*}
\begin{aligned}
    K\cdot \Delta_{min} &\leq tr(P^TD^RP)  \leq K\cdot \Delta_{max}, \\
  K\cdot \Delta_{min} &\leq tr(P^TAP) \leq K\cdot\Delta_{max}.
\end{aligned}
\end{equation*}
Under these conditions, we derive that:
\begin{equation*}
\begin{aligned}
 &\frac{1}{K}\sum_{k=1}^{K}\left\{\frac{P_k^T w^R P_k}{P^T_kD^R P_k} + \frac{P_k^T D P_k}{P_k^T A P_k}\right\} \\
 &\leq  \frac{\sum_{k=1}^{K} P_k^T w^R P_k}{K\cdot \Delta_{min}} + \frac{\sum_{k=1}^{K} P_k^T A P_k}{K\cdot \Delta_{min}}.
 \end{aligned}
\end{equation*}
Then, we obtain the following inequation:
\begin{equation*}
\begin{aligned}
   & \frac{1}{K}\sum_{k=1}^{K}\left\{\frac{P_k^T w^R P_k}{P^T_kD^R P_k} + \frac{P_k^T D P_k}{P_k^T A P_k}\right\} \\
  &  \leq \frac{\Delta_{max}}{\Delta_{min}} \left\{\frac{tr\left(P^T w^R P\right)}{tr\left(P^T D^R P\right)}+\frac{tr\left(P^T D P\right)}{tr\left(P^T A P\right)}\right\}.
    \end{aligned}
\end{equation*}
Therefore, under the condition that $\Delta_{min} \neq 0$, minimizing Eq.~\eqref{eq:10} is equivalent to minimizing the following Eq.~\eqref{eq:11}. This transformation is in line with MinCutPool~\cite{spectralpool}:
\begin{equation}\label{eq:11}
\begin{aligned}
    \underset{P}{\arg\min} &\left\{\frac{tr\left(P^T w^R P\right)}{tr\left(P^T D^R P\right)}+\frac{tr\left(P^T D P\right)}{tr\left(P^T A P\right)}\right\}, \\
    s.t. & \quad P 1_K = 1_N.
\end{aligned}
\end{equation}
The constraint $P\textbf{1}_K=\textbf{1}_N$ ensures each node is assigned to exactly one cluster, thereby maintaining the validity of the partitioning. However, in node clustering, it is possible that a cluster contains only a single node, in which case the minimum value of $\Delta_{min}=0$ under extreme conditions. Such a degenerate scenario would render the optimization of Eq.~\eqref{eq:10} ill-posed. To circumvent this issue, we reformulate Eq.~\eqref{eq:10} into Eq.~\eqref{eq:11}. Moreover, in the design of Eq.~\eqref{eq:9}, we impose a reciprocal constraint between the intra-cluster and inter-cluster degree distributions and their corresponding edge weight distributions. This reciprocal constraint acts as a mutual regularizer that effectively prevents extreme cases from arising.

Notice that problem~\eqref{eq:11} involves graph node clustering based on the principle of dense intra-group connections and sparse inter-group connections. This is a normalized cut problem and is NP-hard. Following the approach of MinCutPool, we employ a deep neural network to approximate the solution by learning a continuous mapping $P$. Technically, we integrate a Graph Convolutional Neural Network (GCNN) and optimize a relaxed version of the loss function, applying a softmax activation to the output layer to obtain a probabilistic assignment. Specifically, we learn a near-optimal soft cluster assignment matrix $P$ by minimizing the following unsupervised loss function (Eq.\eqref{eq:13}):
\begin{align}\label{eq:13}
    \mathcal{L}_u=\frac{tr\left(P^T w^R P\right)}{tr\left(P^T D^R P\right)}+\frac{tr\left(P^T D P\right)}{tr\left(P^T A P\right)}+\left\Arrowvert \frac{P^TP}{\|P^TP\|_F}-\frac{I_K}{\sqrt{K}}\right\Arrowvert_F^2.
\end{align}

By minimizing the loss function $\mathcal{L}_u$, each clique-like structural subgraph is more likely to be grouped into a same cluster depending on the updated edge weight matrix. The learnt assignment matrix $P$ is a real matrix, with $P_{jk}$ close to 1 if node $j$ belongs to cluster $k$. This real matrix gives the opportunity of designing deeper network on graphs and then applying it to enhance the downstream tasks. 

Compared with the MinCutPool, the advantage of our method is that we argue the geometric connection structure between neighbors and leverage the Ollivier-Ricci curvature to measure the connectivity around an edge. The Ricci curvature is very important information especially for extracting node clusters in the graph. 
\subsection{Graph Pooling via Node Clustering}\label{subsec:pool}

We conduct the pooling operation on graph nodes via the clustering assignment matrix $P$ to aggregate the nodes within each cluster into a new supernode. In this way, since the node set is partitioned into $K$ clusters, the new pooled graph $G^{pool}$ contains $K$ supernodes. We need to recalculate two matrices around the pooled graph $G^{pool}$. One is the coarsened adjacency matrix $A^{pool}$, and the other is the feature matrix $X^{pool}$ of the new supernodes. Suppose the graph pooling layer is added behind the $i$-th convolutional layer, then the pooled results are shown as follows:
\begin{align}\label{eq:14}
    A^{pool} = P^T A P; \quad
    X^{pool} = P^T X^{i},
\end{align}
where $A^{pool}\in \mathbb{R}^{K\times K}$ denotes the pooled adjacent matrix as well as a symmetric matrix. $A_{j,j}^{pool}$ indicates the weights sum of all the edges between the nodes in $j$-th cluster, while $A_{j,k}^{pool}$ is the weights sum of the edges between cluster $j$ and $k$. $X^{pool}$ is the corresponding feature matrix, whose entries  $x_{j,k}^{pool}$ in $X^{pool} \in \mathbb{R}^{K\times F}$ are the sum of feature $k$ among the elements in cluster $j$, weighted by the cluster assignment scores.

To avoid the obstacle of self-loops to the propagation in the convolutional layer of GCNN, we compute a normalized adjacency matrix $\Tilde{A}^{pool}$:
\begin{align}
    \hat{A}^{pool} = A^{pool}+I_K ; \quad \Tilde{A}^{pool} = \hat{D}^{-1}\hat{A}^{pool}.
\end{align}
$\hat{A}^{pool}$ is used to add self-loops to the adjacency matrix, and $\hat{D}$ is a diagonal matrix with $\hat{D}_{j,j}= \sum_{k}\hat{A}_{j,k}^{pool}$. The pooled node features are transformed  into the $(i+1)$-th convolutional layer by updating the following operation:
\begin{align}
   X^{i+1} = f(\Tilde{A}^{pool}X^{pool}T^{pool}). 
\end{align}
$T^{pool}$ denotes the corresponding trainable weight matrix to perform the feature transformation followed by Eq.~\eqref{eq:2}.

Overall, in our proposed method, we assume the initial adjacent matrix is $A \in (0,1)^{N\times N}$, with equal weight for each edge. We employ the Ollivier-Ricci flow to iteratively update the edge weight matrix according to the geometric structure of the node connections, facilitating the natural distinction between node clusters. Subsequently, motivated by the spectral clustering method, we learn a new clustering assignment matrix. The coarsened adjacency matrix $A^{pool}$ preserves the global topology structure of the graph, while the cluster embeddings $X^{pool}$ retain the geometric structure of the attribute features from the previous layer $X^i$.  The overall process is illustrated in Fig.~\ref{fig:framework} and Algorithm~\ref{alg:2}.

Our proposed Ricci flow-based clustering method serves as a general technique applicable to solving clustering tasks on any graph-structured data. In this study, we specifically concentrate on graph pooling techniques combined with GCNN to learn high-level representations for graphs. Ricci flow describes the evolution process of non-Euclidean domains and offers strong geometric interpretability. By leveraging the Ollivier-Ricci flow, we can differentiate between clique-like structure subgraphs and tree-like structure subgraphs. These two structures contain embedded semantic information, such as positive or negative node pairs, which can provide additional unsupervised information for downstream tasks and enhance performance.

\begin{algorithm}[tb]
    \caption{RicciPool}
  \label{alg:2}
 \begin{algorithmic}
   \STATE {\bfseries Input:} the initial adjacency matrix $A\in(0,1)^{N\times N}$,\\ 
   \quad \quad \quad the  feature matrix $X^i$ of $i$-th convolutional layer, \\
   \quad \quad \quad the number of clusters $K$ and the iteration value $l$.\\
   \STATE 1. Initialize that $w^0 = A$, $d^0(x,y) = w^0_{xy}$. 
   \STATE 2. Iterate the adjacency matrix $w^0$ by Ollivier-Ricci flow $l$ times following Algorithm~\ref{alg:example}.
   \STATE 3. Train the loss function Eq.~\eqref{eq:13} using GCNN and obtain an assignment matrix $P\in\mathbb{R}^{N\times K}$. 
   \STATE 4. Calculate $A^{pool}$ and $X^{pool}$ after pooling using Eq.~\eqref{eq:14}.
   \STATE {\bfseries Output:} the input adjacency matrix $A^{pool}$ and feature matrix $X^{pool}$ of the $(i+1)$-th convolutional layer.
 \end{algorithmic}
\end{algorithm}

\subsection{Computational Complexity Analysis}

In this section, we delve into the computational complexity of our proposed RicciPool theoretically. As mentioned earlier, assuming the graph has $N$ nodes, $K$ cluster assignments, and an initial feature dimension is $F$, with $N > K$. The total time cost of our GCNN-based pooling method primarily consists of two parts. The first part involves the computational complexity of the Ollivier-Ricci flow. Let's assume there are $l$ iterations in our Ollivier-Ricci flow operation. In each iteration, computing the Ollivier-Ricci curvature takes a time cost of $\mathcal{O}(N^4\log^2N)$~\cite{siddharth2021}, hence the total computational complexity of the Ollivier-Ricci flow is $\mathcal{O}(lN^4\log^2N)$. The second part deals with the computational complexity of GCNN. The time cost of one GCNN layer is quantified by $\mathcal{O}(N^3 + N^2 F +NF^2)$ which is approximately  $ \mathcal{O}(N^3)$. 
Similar to~\cite{structpool},let's suppose the number of iterations in our training network is $m$, then the total computational cost of this part is $\mathcal{O}(mN^3)$. After the pooling operation, the final step involves computing the pooled adjacent matrix $A^{pool}$ and the associated feature matrix $X^{pool}$, which takes about $\mathcal{O}(NKF + N^2K + NK^2)$ approximately $ \mathcal{O}(N^3)$. Altogether, the total computational complexity of our GCNN-based RicciPool is $\mathcal{O}((lN\log^2N+m)N^3)$. 

Compared with other GCNN based graph pooling methods, RicciPool uses the Ollivier-Ricci flow to update the edge weight matrix before network training, without changing the architecture, training process, or inference process of the GCNN itself. Therefore, the curvature computation should be regarded as an offline preprocessing cost in the full pipeline, rather than an additional cost inside GCNN training or testing. As discussed in the scalability analysis (detailed in Section~\ref{sec:large_graph_analysis} and Section~\ref{sec:cost_anslysis}), this cost is more manageable on sparse or moderately connected graphs, while it may become expensive on highly dense graphs.

\begin{table*}
\vspace{-0cm}
	\caption{Dataset statistics. }
	\vspace{-0.2cm}
	\label{table:1}
	\setlength\tabcolsep{4pt} 
	\begin{center}
			\begin{sc}
				\begin{tabular}{lccccc}
					\toprule
					name & Category & Nodes (avg)& Edges (avg) & Graphs & Classes\\
					\midrule
					ENZYMES    & Bioinformatics & 32.63& 124.20 & 600 & 6  \\
					PROTEINS & Bioinformatics & 39.06 & 72.82 & 1113 & 2\\
					D\&D & Bioinformatics & 284.32 & 1431.3 & 1178 & 2 \\
					\midrule
					IMDB-B     & Social Network & 19.77 & 96.53& 1000& 2\\
					IMDB-M      & Social Network & 13.00 & 65.94 &1500 & 3\\
           COLLAB    & Social Network & 74.00 & 2457&5000 & 3\\
            REDDIT-M-5K   & Social Network & 508.52 & 594.87 &5000 & 5\\
					\bottomrule
				\end{tabular}
			\end{sc}
	\end{center}
	\vspace{-0.05cm}
\end{table*}

\begin{table*}[htbp]
\centering
\caption{Comparison results between different graph pooling methods under the same framework with the same number of clusters $K = 32$ and the same Ollivier-Ricci flow iteration value $l = 5$ for RicciPool. All the comparison results represent ten-run averages.}
\label{table:2}
\setlength\tabcolsep{6pt}
\begin{sc}
\begin{tabular}{cc*{7}{c}}  
\toprule
\multirow{2}{*}{\textbf{Category}} & \multirow{2}{*}{\textbf{Method}} & \multicolumn{7}{c}{\textbf{Dataset}} \\
\cmidrule(lr){3-9}
& & \textbf{Enzymes} & \textbf{D\&D} & \textbf{Proteins} & \textbf{IMDB-B} & \textbf{IMDB-M} & \textbf{Collab} & \textbf{REDDIT-M-5K} \\
\midrule
\multirow{6}{*}{\makecell[c]{Node \\Selection\\Pooling}} 
& SumPool~\cite{selfpool} & 41.33$\pm$5.31 & 77.26$\pm$3.51 & 77.21$\pm$4.34 & 56.90$\pm$3.01 & 39.80$\pm$3.20 & 73.46$\pm$1.82 & 49.15$\pm$2.58 \\
& SortPool~\cite{zhang2018graphclassification}& 34.67$\pm$3.71 & 76.75$\pm$3.10 & 73.96$\pm$4.46 & 60.70$\pm$3.55 & 39.87$\pm$3.51 & 71.04$\pm$2.03 & 49.63$\pm$1.55 \\
& TopkPool~\cite{gao2019} & 31.67$\pm$3.94 & 72.99$\pm$3.83 & 73.96$\pm$4.33 & 61.30$\pm$2.53 & 41.53$\pm$3.13 & 71.22$\pm$1.62 & 48.23$\pm$2.08 \\
& SagPool~\cite{selfpool} & 32.00$\pm$3.56 & 74.62$\pm$2.84 & 73.87$\pm$4.50 & 62.80$\pm$4.33 & 42.13$\pm$3.56 & 73.58$\pm$1.82 & 49.61$\pm$1.67 \\
& CCP-GNN~\cite{CCP-GNN} & 52.83$\pm$3.66 & 71.62$\pm$4.47 & 71.89$\pm$5.33 & 73.80$\pm$2.86 & 50.80$\pm$2.81 & 78.98$\pm$2.13 & 53.39$\pm$2.98 \\
& MID~\cite{MID} & 41.05$\pm$1.88 & 77.56$\pm$4.46 & 75.05$\pm$3.18 & 73.64$\pm$5.53 & 51.47$\pm$3.22 & 80.19$\pm$2.39 & \textbf{56.04$\pm$2.87} \\
\midrule
\multirow{4}{*}{\makecell[c]{Node \\Clustering\\Pooling}} 
& StructPool~\cite{structpool} & 34.67$\pm$5.52 & 76.32$\pm$2.57 & 73.24$\pm$5.51 & 67.40$\pm$3.44 & 46.47$\pm$2.75 & 70.44$\pm$6.65 & 48.07$\pm$1.83 \\
& DiffPool~\cite{diffpool} & 38.67$\pm$6.23 & 73.24$\pm$5.51 & 76.94$\pm$4.36 & 55.80$\pm$2.96 & 39.20$\pm$2.92 & 73.26$\pm$2.23 & 50.15$\pm$2.02 \\
& MinCutPool~\cite{spectralpool} & 40.67$\pm$5.33 & 78.89$\pm$2.89 & 76.85$\pm$4.34 & 58.00$\pm$5.25 & 40.60$\pm$4.35 & 73.32$\pm$1.98 & 50.07$\pm$1.75 \\
& SEP~\cite{SEP} & 37.00$\pm$9.36 & 77.35$\pm$4.94 & 75.50$\pm$5.42 & 73.60$\pm$4.25 & 51.53$\pm$3.51 & \textbf{81.40$\pm$1.33} & 49.99$\pm$1.71 \\
\midrule
\multirow{4}{*}{\makecell[c]{Geometry \\Aware\\Pooling}} 
& ORC~\cite{orc_arxiv} & 38.50$\pm$8.45 & 81.62$\pm$2.14 & 79.01$\pm$3.37 & 64.50$\pm$5.33 & 43.07$\pm$3.59 & 73.80$\pm$1.98 & 51.53$\pm$1.82 \\
& SpreadPool~\cite{GAEP} & 66.30$\pm$7.60 & 80.60$\pm$2.21 & 76.40$\pm$3.03 & 75.80$\pm$4.07 & 50.40$\pm$2.85 & 67.72$\pm$2.94 & - \\
& MagPool~\cite{GAEP} & \textbf{67.50$\pm$4.55} & 80.17$\pm$2.81 & 76.04$\pm$2.83 & 75.10$\pm$3.45 & 50.60$\pm$2.84 & 67.84$\pm$2.87 & - \\
& \textbf{RicciPool} & 62.42$\pm$6.46 & \textbf{82.99$\pm$3.23} & \textbf{79.46$\pm$3.82} & \textbf{76.00$\pm$4.53} & \textbf{51.62$\pm$3.42} & 75.30$\pm$1.73 & 55.20$\pm$2.02 \\
\bottomrule
\end{tabular}
\end{sc}
\end{table*}

\section{Experiments}\label{sec:exp}
\subsection{Datasets and Experimental Settings}
\label{exp:datasets}
We evaluate proposed RicciPool on seven benchmark datasets, including three bioinformatics datasets: ENZYMES, PROTEINS~\cite{Karsten2005}, D\&D~\cite{D&D2003}; four social networks:  IMDB-B~\cite{Pinar2015a}, IMDB-M~\cite{Pinar2015b}, COLLAB~\cite{collab}, and REDDIT-M-5K~\cite{reddit-multi-5k}.  The statistics and properties of these datasets are reported in Table~\ref{table:1}, following the introduction in~\cite{structpool}. Almost all existing graph pooling methods run on these datasets, as they are relatively large-scale with a significant number of graphs.

We compare our method with three categories of state-of-the-art approaches: 1) Node selection-based methods, including SUMPool~\cite{selfpool}, SORTPool~\cite{zhang2018graphclassification}, TOPKPool~\cite{gao2019}, SAGPool~\cite{selfpool}, MID~\cite{MID}, and CCP-GNN~\cite{CCP-GNN};
2) Node clustering-based methods, including DIFFPool~\cite{diffpool}, StructPool~\cite{structpool}, MinCutPool~\cite{spectralpool}, and SEP~\cite{SEP};
3) Geometry-aware methods. Since our method also falls into the geometry-aware category, we further include comparisons with ORC~\cite{orc_arxiv}, SPREADPool~\cite{GAEP}, and MAGPool~\cite{GAEP} from this category.

Our method was implemented in Python 3.8 and experiments were conducted on a server configured with an Inter(R) Xeon(R) Gold 6246R CPU, 256GB RAM and four NVIDIA GeForce RTX 3090 GPUs. Within this setup, the curvature precomputation was handled by the CPU, while the graph pooling model was trained on the GPUs. Ollivier-Ricci flow computed using open-source tool GraphRicciCurvature\footnote{https://github.com/saibalmars/GraphRicciCurvature.}. The code and experiments are available at \url{ https://github.com/cqfei/RicciPool/tree/master}.

\vspace{-0.3cm}
\subsection{Classification Results}
To demonstrate the effectiveness of our proposed RicciPool, we compare our method with $13$ graph pooling methods on the seven standard datasets shown in Table~\ref{table:1} within the same network framework, following the protocol outlined in~\cite{structpool}. All pooling methods are applied in the spatial GCNN framework introduced in Section~\ref{GCNN}. The comparison results are presented in Table~\ref{table:2}, with the best results highlighted in bold. 
Furthermore, for all methods, we conducted $10$-fold cross-validations and computed the average accuracy and standard deviation of each dataset. In our RicciPool, the Ollivier-Ricci flow iteration steps $l$ are set to $5$. The hyperparameters $\alpha=0.5$ and $\eta=1.0$ are fixed adhere to the recommendations of~\cite{ni2019community},  consistent with all other experiments in this paper. An ablation analysis of these hyperparameters is detailed in Section~\ref{ablation}. To ensure a fair comparison, we re-implemented all baseline methods using the same data split and uniform hyperparameters (learning rate: $1e-3$; hidden dimension: $64$; maximum training epochs: $150$). It should be noted that, due to the high time complexity of the SPREADPOOL and MAGPOOL methods, we were unable to complete their computations on the REDDIT-M-5K dataset within a feasible timeframe using the same computing platform as the other baseline methods, therefore, the corresponding results are not available (marked as ‘-’).

As shown in Table~\ref{table:2}, RicciPool achieves the best performance among the compared methods on D\&D, PROTEINS, IMDB-B, and IMDB-M, which demonstrates the effectiveness of our proposed curvature flow pooling strategy.
To illustrate the role of the curvature flow, we focus on comparing MinCutPool and StructPool. While these two methods primarily rely on pairwise connections between neighboring nodes, RicciPool can distinguish different substructures based on the Ollivier-Ricci curvature flow, thereby offering stronger geometric interpretability. The consistent improvements over these two pooling methods on all seven datasets suggest that curvature flow provides useful structural information for graph pooling.

Furthermore, in comparisons with existing geometry-aware methods, RicciPool achieves better results than existing geometry-aware methods on most datasets, except on ENZYMES. This suggests that refining the node assignment matrix via curvature flow can be beneficial for graph pooling. 
On the ENZYMES dataset, RicciPool does not achieve the best result among all compared methods, but it still outperforms representative clustering based pooling methods such as MinCutPool and StructPool. This shows that curvature flow remains useful, although its additional gain is not sufficient to surpass the strongest geometry based methods.

On the COLLAB dataset, RicciPool also fails to show a significant improvement. The reason is that this is a highly dense graph (average density $\approx 0.91$) where the Ollivier-Ricci curvature of most edges is positive. This causes the geometric flow to adjust edge weights in a consistent direction, weakening its ability to discriminate between graph structures and limiting the refinement effect of the curvature flow on the node assignment matrix. Unlike COLLAB, REDDIT-M-5K is extremely sparse and close to a tree like structure (average density $\approx 0.0046$). In this case, many edges tend to have negative curvature, and the Ricci flow also adjusts edge weights in a consistent direction as opposed to positive curvature, which limits its discriminative effect. These results indicate that both overly dense and overly sparse graphs may weaken the effectiveness of RicciPool. Overall, RicciPool is more suitable for graphs with moderate connectivity and distinct cluster structures of nodes.

\begin{table*}
	\caption{Comparison results with the baseline which excludes the graph pooling layer. For RicciPool, the number of clusters is set to $K = 32$ and the Ollivier-Ricci flow iteration value is set to $l = 5$. All the comparison results represent ten-run averages.}
	\label{table:3}
	\setlength\tabcolsep{6pt} 
	\begin{center}
			\begin{sc}
				\begin{tabular}{p{1.6cm}ccccccc}
					\toprule
					\textbf{Dataset}& ENZYMES & D\&D & Proteins &IMDB-B &IMDB-M & COLLAB& REDDIT-M-5K \\
					\midrule
                    No Pooling  & 57.33 & 81.42  & 78.27 & 73.40 & 50.53 & 75.26 &54.6\\
					RicciPool & \textbf{62.42} &\textbf{82.99} & \textbf{79.46} &\textbf{76.00} &\textbf{51.62} & \textbf{75.30} &\textbf{55.2}\\
					\bottomrule
				\end{tabular}
			\end{sc}
	\end{center}
    \vspace{-0.5cm}
\end{table*}

\subsection{Ablation Study and Analysis}
\label{ablation}
In the following, we evaluate RicciPool under various parameter settings and provide a detailed discussion based on a series of comparison experiments.

\textbf{Effects of graph pooling layer.} We conduct an ablation analysis on the graph pooling layer to demonstrate its impact on the final graph classification results. Through a set of experiments, we compare our proposed RicciPool with a baseline method lacking the graph pooling layer, and present the results in Table~\ref{table:3}. From Table~\ref{table:3},  it's evident that without the graph pooling layer, the baseline method behaves similarly to traditional GCNNs. Like RicciPool, the baseline method adheres to the network framework described in Section~\ref{GCNN} and undergoes 10-fold cross-validation to compute the average accuracy for each graph dataset. Notably, our proposed method outperforms the baseline across all $7$ datasets,
indicating that the graph pooling operation contributes to 
extracting useful features among nodes and facilitating effective graph-level embedding representation. 

\begin{figure}
	\centering
	\includegraphics[width=1.0\linewidth]{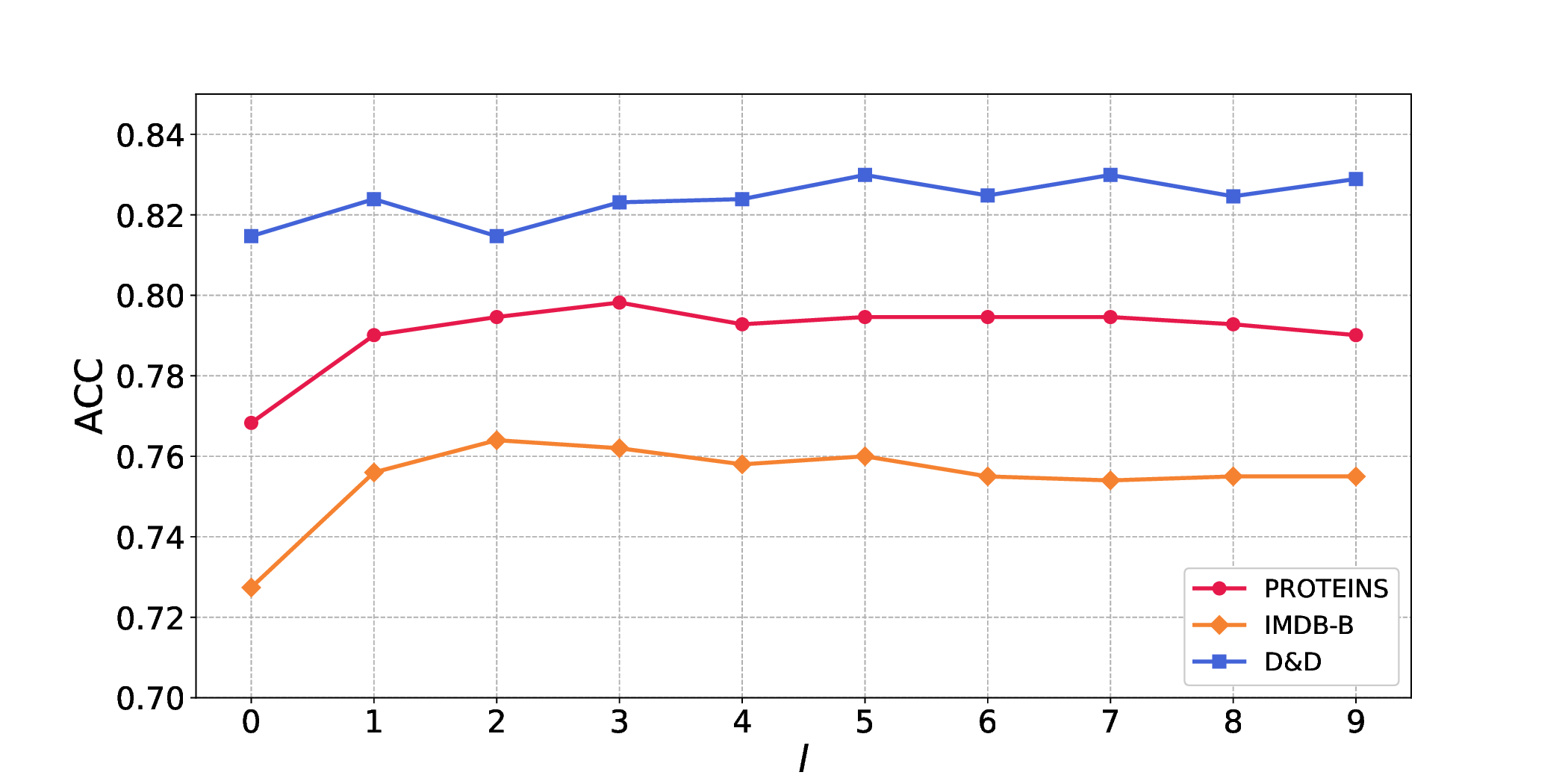}
	\caption{The prediction accuracy of RicciPool under different Ollivier-Ricci flow iterations $l$ with the same number of clusters $K = 32$.}
	\label{fig:iter}
\end{figure}

\textbf{Effects of Ricci flow iterations $l$.} We analyze how the number of Ricci flow iterations $l$ affects prediction performance on D\&D, PROTEINS, and IMDB-B, with the results shown in Table~\ref{table:4} and Fig.~\ref{fig:iter}. Compared with $l=0$, where no Ricci flow based edge reweighting is applied, using $l>0$ generally improves classification performance, indicating that Ollivier Ricci flow provides useful local structural information for graph pooling. As $l$ increases, positive curvature edges are gradually shortened and negative curvature edges are stretched, making dense substructures more compact and sparse or bridge like connections easier to separate. However, the optimal iteration number varies across datasets, with $l=5$ for D\&D, $l=3$ for PROTEINS, and $l=2$ for IMDB-B. This suggests that different datasets require different degrees of curvature flow refinement. However, excessive iterations may over adjust edge weights and weaken useful topology for node assignment learning.

\begin{table}
	\caption{The prediction accuracy of RicciPool under different Ollivier-Ricci flow iterations $l$ with the same number of clusters $K = 32$. }
	\label{table:4}
	\setlength\tabcolsep{6pt} 
	\begin{center}
			\begin{sc}
				\begin{tabular}{p{1.2cm}cccccc}
					\toprule\multirow{2}{1.5cm}{Dataset} & \multicolumn{6}{c}
				{Iteration} \\\cmidrule(lr){2-7}
					&$l = 0$& $l = 1$& $l=2$ & $l=3$ & $l=4$ & $l=5$  \\
					\midrule
					D\&D & 81.47 & 82.39 & 81.47& 82.31 & 82.39 &\textbf{82.99}\\
					Proteins & 76.83 & 79.01  &79.46 & \textbf{79.82} & 79.28 &79.46\\
					IMDB-B  &72.74 & 75.60 &\textbf{76.40}& 76.20 & 75.80 & 76.00\\
					\bottomrule
				\end{tabular}
			\end{sc}
	\end{center}
\end{table}

\begin{table}
	\caption{Curvature convergence statistic of the Ollivier-Ricci flow.}
	\label{table:Curvatureconvergence}
	\setlength\tabcolsep{6pt} 
	\begin{center}
			\begin{sc}
				\begin{tabular}{lccccc}
					\toprule
					\textbf{Dataset}& ENZYMES & Proteins &IMDB-B &IMDB-M\\
					\midrule
                       \makecell{Expectation} & 11 & 10  & 6 & 6 \\
					Proportion & 60.67 &59.03 & 72.6 &43.53 \\
					\bottomrule
				\end{tabular}
			\end{sc}
	\end{center}
\end{table}

To further explain this phenomenon, we further investigate the correlation between curvature convergence and optimal performance in RicciPool. We analyzed the convergence properties of Ollivier-Ricci flow on four benchmark datasets, with convergence defined as a difference between maximum and minimum edge curvature below $10^{-3}$. 
Statistical results for curvature convergence iterations are summarized in Table~\ref{table:Curvatureconvergence}. 
Two trends can be observed.

First, larger graphs usually require more iterations to reach curvature convergence.  
For example, PROTEINS needs about $10$ iterations, compared with about $6$ for IMDB-B.
This is because larger graphs often contain more diverse local connection patterns and more types of edges, such as dense intra community edges, sparse branch edges, and bridge edges. These different edge types evolve at different speeds under Ricci flow, so more iterations are needed for the curvature values and the induced edge weights to stabilize.

Second, datasets with slower curvature convergence tend to require more RicciPool iterations to reach the best classification performance. 
For example, PROTEINS converges more slowly and peaks at $l=3$, whereas IMDB-B converges faster and peaks at $l=2$.
This suggests that curvature convergence speed reflects the complexity of geometric regularization. 

Third, the optimal iteration number for classification is usually smaller than the full convergence iteration number. 
For example, after PROTEINS peaks at $l=3$ and IMDB-B peaks at $l=2$, further increasing $l$ does not bring additional gains.
RicciPool only needs moderate curvature evolution to provide sufficient structural separation for node clustering. Excessive iterations may over contract positive curvature edges and over stretch negative curvature edges, which can distort useful graph connectivity and accelerate over squashing and over smoothing during subsequent GNN propagation. Therefore, the iteration number should balance structural discrimination and graph connectivity preservation.

\begin{figure}
	\centering
	\includegraphics[width=0.48\linewidth]{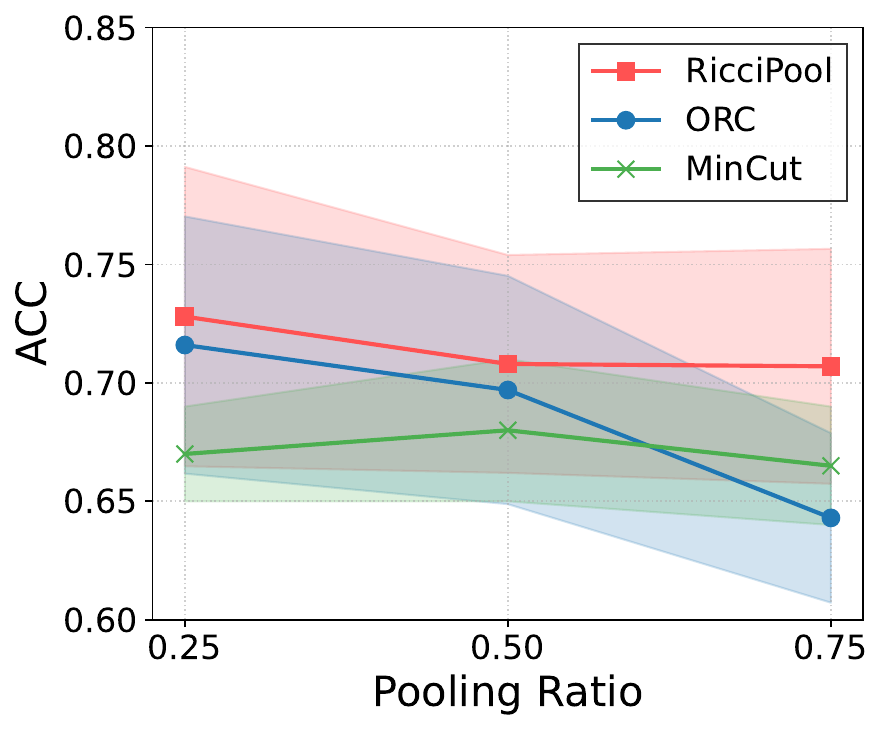}
    \includegraphics[width=0.48\linewidth]{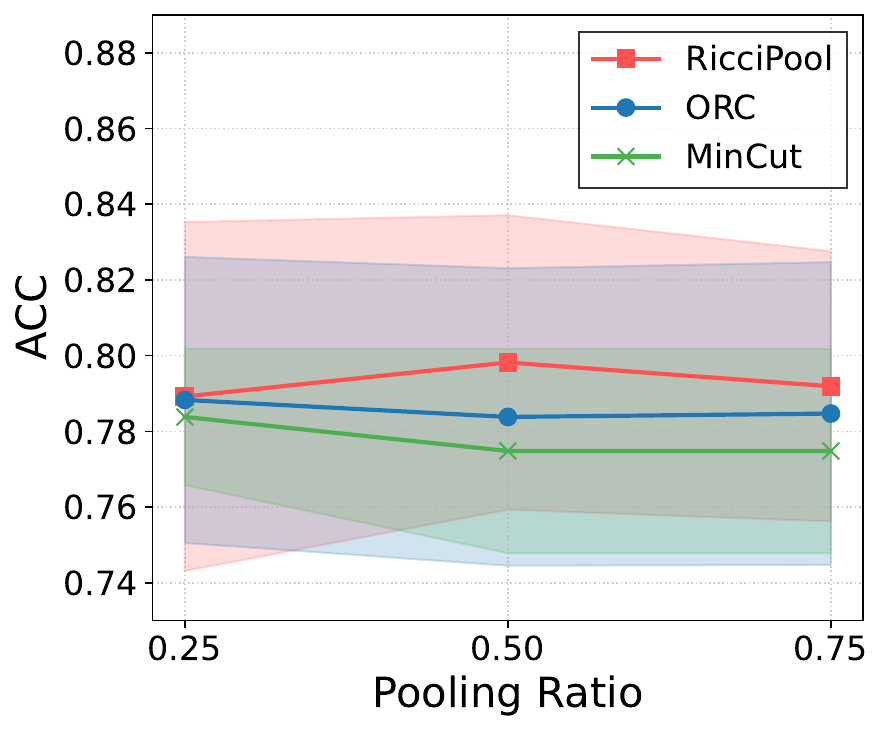}
	\caption{Comparison under different pooling rates with the same Ollivier-Ricci flow iteration $l=5$ (left: IMDB-B, right: PROTEINS).}
	\label{fig:polling_ratio_cmp}
    \vspace{-0.5cm}
\end{figure}
\textbf{Effects of graph pooling rate $r$.} In this segment, we examine how the number of clusters $K$ influences RicciPool by setting a pooling rate $r \in (0,1)$ to control $K$ ($r=1$ means $K=64$). We adjust $r$ to $\{0.25, 0.50, 0.75\}$ to evaluate the sensitivity of graph pooling methods to the pooling ratio. Figure~\ref{fig:polling_ratio_cmp} compares the performance of RicciPool, ORC (also a Ricci Flow-based method), and MinCutPool under different pooling ratios. The results show that RicciPool maintains the highest average classification accuracy across all ratios, validating the effectiveness of its core mechanism. It is worth noting that RicciPool exhibits a relatively wide performance distribution, which reflects the sensitivity of its curvature-based global reweighting mechanism to graph structural features.
Further comparison reveals that ORC, another geometry-aware method, outperforms MinCut on average. However, ORC shows a larger performance decrease at the high pooling ratio ($0.75$) and exhibits higher variability than RicciPool in our experiments, which may suggest that it is more sensitive to structural compression under these settings.
In contrast, MinCut shows narrower confidence intervals and more stable results, but its average performance is lower than RicciPool. This suggests that MinCut is stable but may be less effective in extracting discriminative structural information under different pooling ratios.
In summary, despite some performance variability, 
RicciPool achieves competitive and generally better performance across different pooling ratios, supporting  the advantage of its curvature-driven mechanism in extracting key structural information. Compared to ORC, RicciPool shows less performance degradation at higher compression ratios in our experiments.

\begin{figure}[ht]
    \centering
        \begin{subfigure}[b]{0.24\textwidth}
        \centering
        \includegraphics[width=\textwidth]{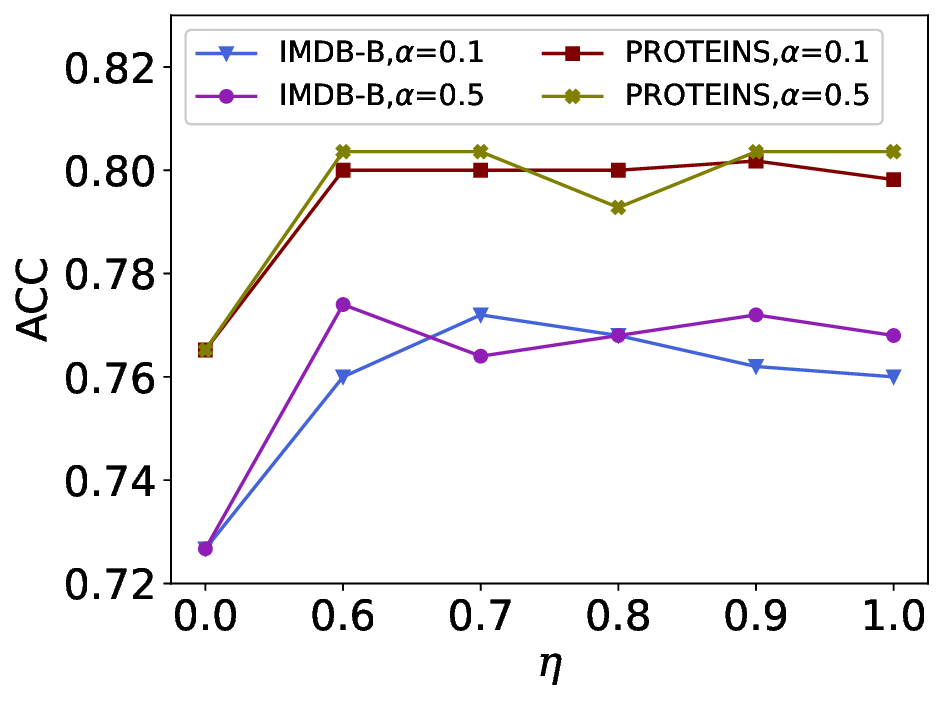}
        \caption{}
        \label{fig:lr}
    \end{subfigure}
    \hfill
    \begin{subfigure}[b]{0.24\textwidth}
        \centering
        \includegraphics[width=\textwidth]{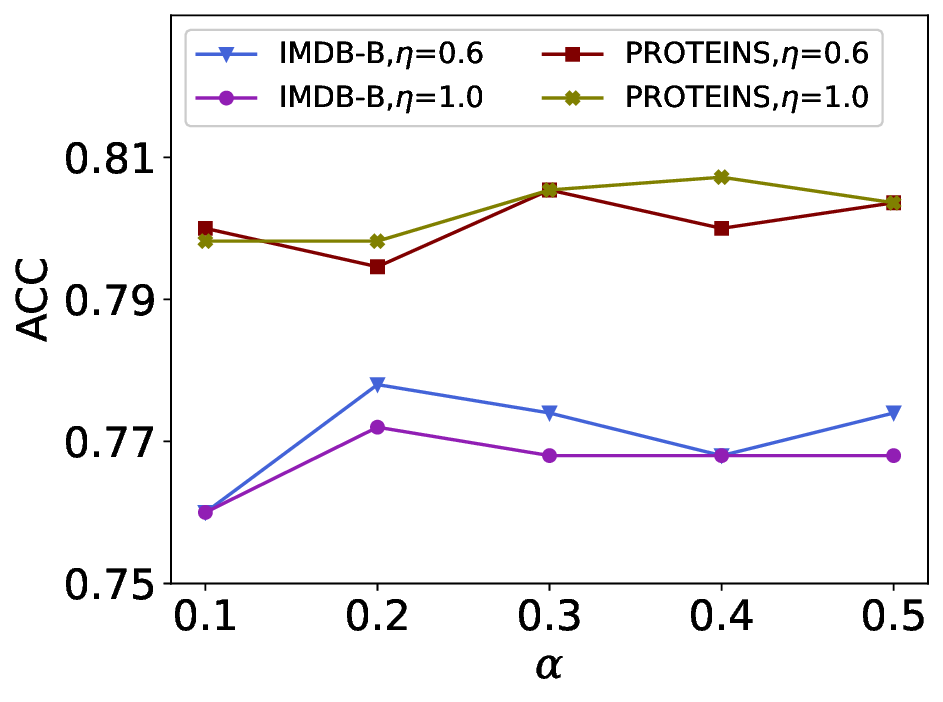}
        \caption{}
        \label{fig:alpha}
    \end{subfigure}
    \caption{Performance on different $\eta$ (left) and $\alpha$ (right) values for dataset IMDB-B and PROTEINS.}
    \label{fig:eta_and_alpha}
\end{figure}

\textbf{Effects of curvature intervention and neighbor information.} We introduced the hyperparameter $\eta$ (ranging from $0$ to $1$) to observe the impact of curvature information on our algorithm. The experimental results are shown in Fig.~\ref{fig:lr}. It can be seen that for the IMDB-B dataset, our method achieved optimal performance at $\eta=0.7$ and $\eta=0.6$ when $\alpha=0.1$ and $\alpha=0.5$, respectively. For the PROTEINS dataset, the best performance was achieved at $\eta=0.9$ and $\eta=1.0$. Based on this, we draw two conclusions: 1) Incorporating curvature information is indeed effective ($\eta=0$ represents results without curvature information); 2) The optimal level of curvature intervention ($\eta$) differs between the IMDB-B and PROTEINS datasets. This suggests that the impact of curvature information may be dataset-dependent, and tuning this parameter could be important for optimal performance on a given dataset.
One influencing factor, according to our analysis, is the connectivity of the dataset itself. Structural analysis shows that the IMDB-B dataset is denser than the PROTEINS dataset (the average degree of IMDB-B is higher than that of PROTEINS). Consequently, optimal results for IMDB-B were obtained with less curvature information ($\eta=0.7$ and $\eta=0.6$), whereas the PROTEINS dataset required more curvature intervention ($\eta=0.9$ and $\eta=1.0$) to achieve the best results

Additionally, when computing graph curvature, the hyperparameter $\alpha$ controls the weight distribution between the source and target nodes' probability distributions in the calculation of Wasserstein distance. A smaller $\alpha$ means that the source node’s neighborhood has a greater influence on the calculation, while the target node’s neighborhood has a lesser influence. The experiments regarding this hyperparameter are shown in Fig.~\ref{fig:alpha}. From Fig.~\ref{fig:alpha}, it can be observed that for the IMDB-B dataset, the best performance was achieved at $\alpha=0.2$ for both $\eta=0.6$ and $\eta=1.0$. For the PROTEINS dataset, the optimal results were obtained at $\alpha=0.3$ and  $\alpha=0.4$. Based on this, we draw another conclusion: the influence of neighborhood information varies across different datasets. Both the source and target nodes' neighborhood information affect edge curvature and indirectly influence the final performance of our algorithm.

\textbf{A trick for hyperparameter tuning.}
In the RicciPool method, we introduced three novel Ricci flow-related hyperparameters. Based on extensive experiments, we concluded that optimal performance
is achieved when $\alpha$ ranges from $0.1$ to $0.5$,  $\eta$  ranges from $0.5$ to $1.0$, and iterations
their optimal ranges: $\alpha \in [0.1, 0.5]$, $\eta \in [0.5, 1.0]$, and iteration times generally not exceeding $5$.

\begin{figure*}[htbp]
    \centering
    \begin{subfigure}[a-1]{0.3\linewidth}
        \centering
        \includegraphics[width=0.9\linewidth]{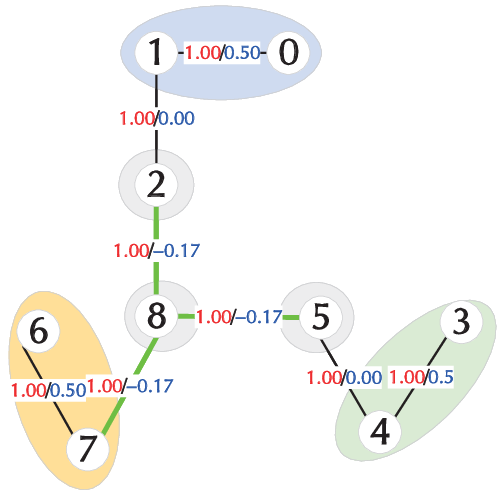}
        \caption{$l=0$}
    \end{subfigure}
    \vspace{5mm}
    \centering
    \begin{subfigure}[a-2]{0.3\linewidth}
     \vspace{5mm}
        \centering    
        \includegraphics[width=0.9\linewidth]{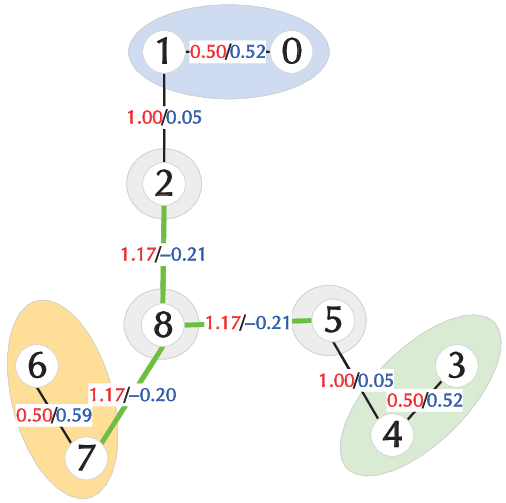}
        \caption{$l=1$}
    \end{subfigure}
    \centering
    \begin{subfigure}[a-3]{0.3\linewidth}
        \centering
        \includegraphics[width=0.9\linewidth]{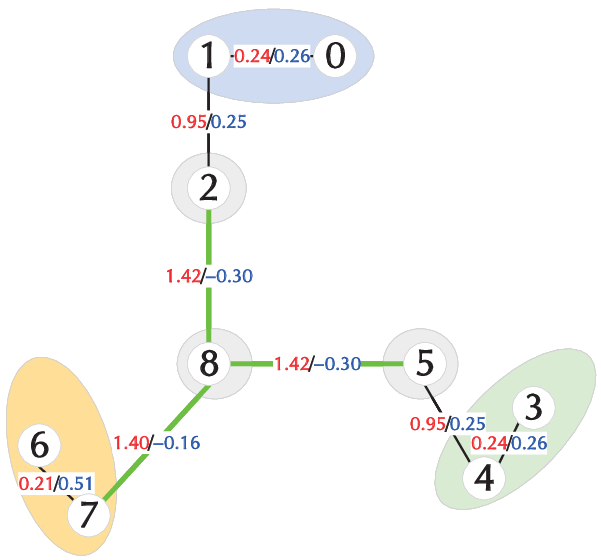}
        \caption{$l=2$}
    \end{subfigure}
\\
    \begin{subfigure}{0.3\linewidth}
        \centering
        \includegraphics[width=0.9\linewidth]{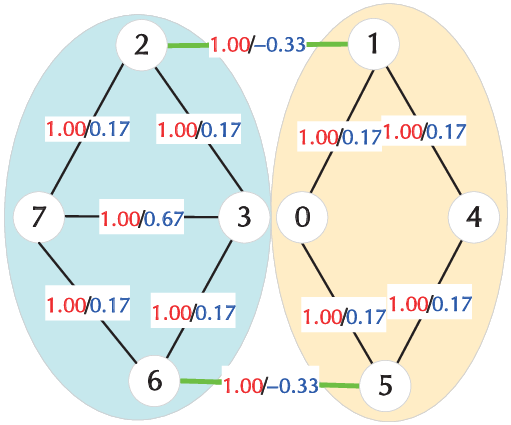}
        \caption{$l=0$}
    \end{subfigure}
    \vspace{5mm}
    \centering
    \begin{subfigure}{0.3\linewidth}
     \vspace{5mm}
        \centering    
        \includegraphics[width=0.9\linewidth]{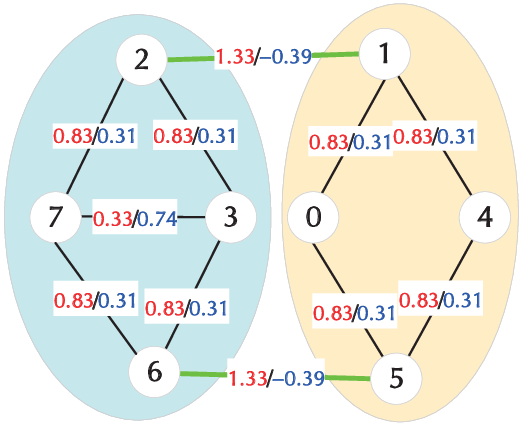}
        \caption{$l=1$}
    \end{subfigure}
    \begin{subfigure}{0.3\linewidth}
    \vspace{5mm}
        \centering
        \includegraphics[width=0.9\linewidth]{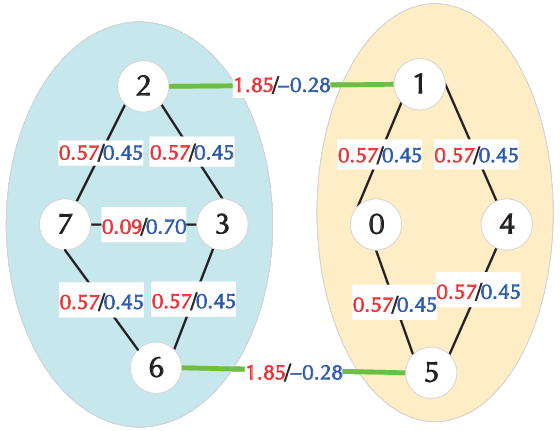}
        \caption{$l=2$}
    \end{subfigure}
    \\
        \begin{subfigure}{0.3\linewidth}
        \centering
        \includegraphics[width=0.9\linewidth]{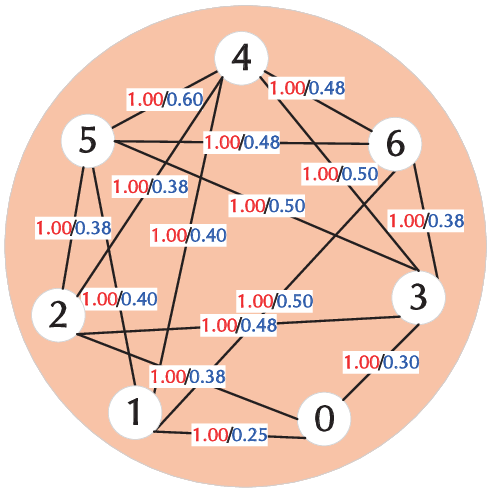}
        \caption{$l=0$}
    \end{subfigure}
    \vspace{5mm}
    \centering
    \begin{subfigure}{0.3\linewidth}
     \vspace{5mm}
        \centering    
        \includegraphics[width=0.9\linewidth]{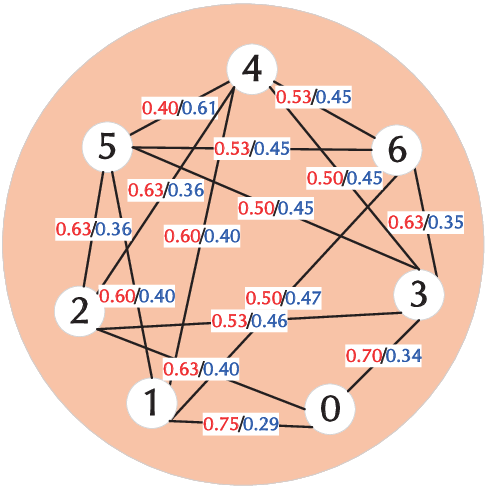}
        \caption{$l=1$}
    \end{subfigure}
    \begin{subfigure}{0.3\linewidth}
    \vspace{5mm}
        \centering
        \includegraphics[width=0.9\linewidth]{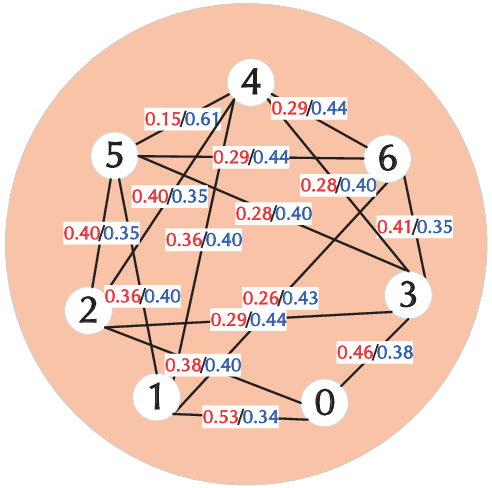}
        \caption{$l=2$}
    \end{subfigure}
   \caption{Visualization of the Ollivier-Ricci flow in RicciPool, using three real example graphs from the ENZYMES (first row, ID $351$, referred to as graph A below) and PROTEINS datasets (IDs $1084$ and $412$, referred to as graphs B and C below). Figures a-c, d-f, and g-i show the states of graphs A, B, and C after $0$, $1$, and $2$ iterations of the Ollivier-Ricci curvature flow, with the numbers on the edges indicating the distances between nodes. The subgraphs with different background colors represent different communities. The lengths of the edges represent the edge weights. The black lines on the edges indicate positive curvature, while the green lines indicate negative curvature.}
   \label{fig:r1}
\end{figure*}

\vspace{-0.4cm}
\subsection{Analysis of Cluster Assignment in RicciPool}
\textbf{Visualization of graphs in RicciPool.} We examined the extent to which RicciPool learns meaningful node clusters by visualizing the node distribution after the Ollivier-Ricci flow, similar to~\cite{diffpool}. Fig.~\ref{fig:r1} illustrates this visualization of node assignments in the RicciPool layer on three graphs from the ENZYMES and PROTEINS datasets. 
Notably, we observe distinct node cluster memberships across different datasets. RicciPool effectively exploits graph curvature to uncover embedded cluster structures, shrinking the edge weight of fully connected subgraphs and extending the edge distance of tree-structured subgraphs. Graphs A, B, and C represent three types of graphs with varying connection strengths: Graph A has a tree-like structure, Graph C has a nearly fully connected structure, and the connectivity of Graph B is intermediate between A and C. We can observe the different changes in the subgraph structures of these graphs after the Ollivier-Ricci flow.

\textbf{Subgraph structures.} 
We illustrate the edge lengths corresponding to the edge weights. Fig.~\ref{fig:r1} depicts the evolving node distributions of the graphs under the Ollivier-Ricci flow. The visualization demonstrates that Ollivier-Ricci curvature effectively measures pairwise connections between neighboring nodes and the density of subgraphs. The flow transforms dense subgraphs into even denser ones (see graphs A and B), while extending the edge lengths of sparse subgraphs. Moreover, Ollivier-Ricci curvature distinguishes various subgraph structures, including fully connected (see graph C) and tree-like formations (see graph A). For instance, in Fig.~\ref{fig:r1} (g-i), graph C exhibits quasi-full connectivity, with node distances decreasing post Ollivier-Ricci flow evolution, thereby tightening connections. These findings underscore the strong geometric nature and interpretability of Ollivier-Ricci curvature.

\textbf{Sensitivity of the Number of Clusters.} \label{sec:analysis}
We observed that cluster assignment varies across different graph datasets and with $K$, the number of clusters, indicating sensitivity to the pooling result. For the Proteins dataset, the optimal $K$ varies depending on the specific graphs. Setting $K$ too large may fragment densely connected subgraphs into separate clusters, leading to a loss of closely connected edges between neighboring nodes. Conversely, if $K$ is too small, the resulting pooled graph may have excessively small node sizes, failing to preserve the global topology of the original graph.
In the case of the IMDB-M dataset, where the average number of nodes per graph is $13$ and the average number of edges is $66$, nearly all graphs are fully connected. Setting $K$ too large can disrupt the fully connected structure. This explains why the IMDB-M dataset shows poorer classification results under graph pooling when $K>1$, compared to the other four datasets presented in Table~\ref{table:2}.
\vspace{-0.3cm}
\subsection{Scalability and Efficiency of RicciPool on Large-Scale Graph Datasets}
\label{sec:large_graph_analysis}
\begin{figure}
	\centering
	\includegraphics[width=1.0\linewidth]{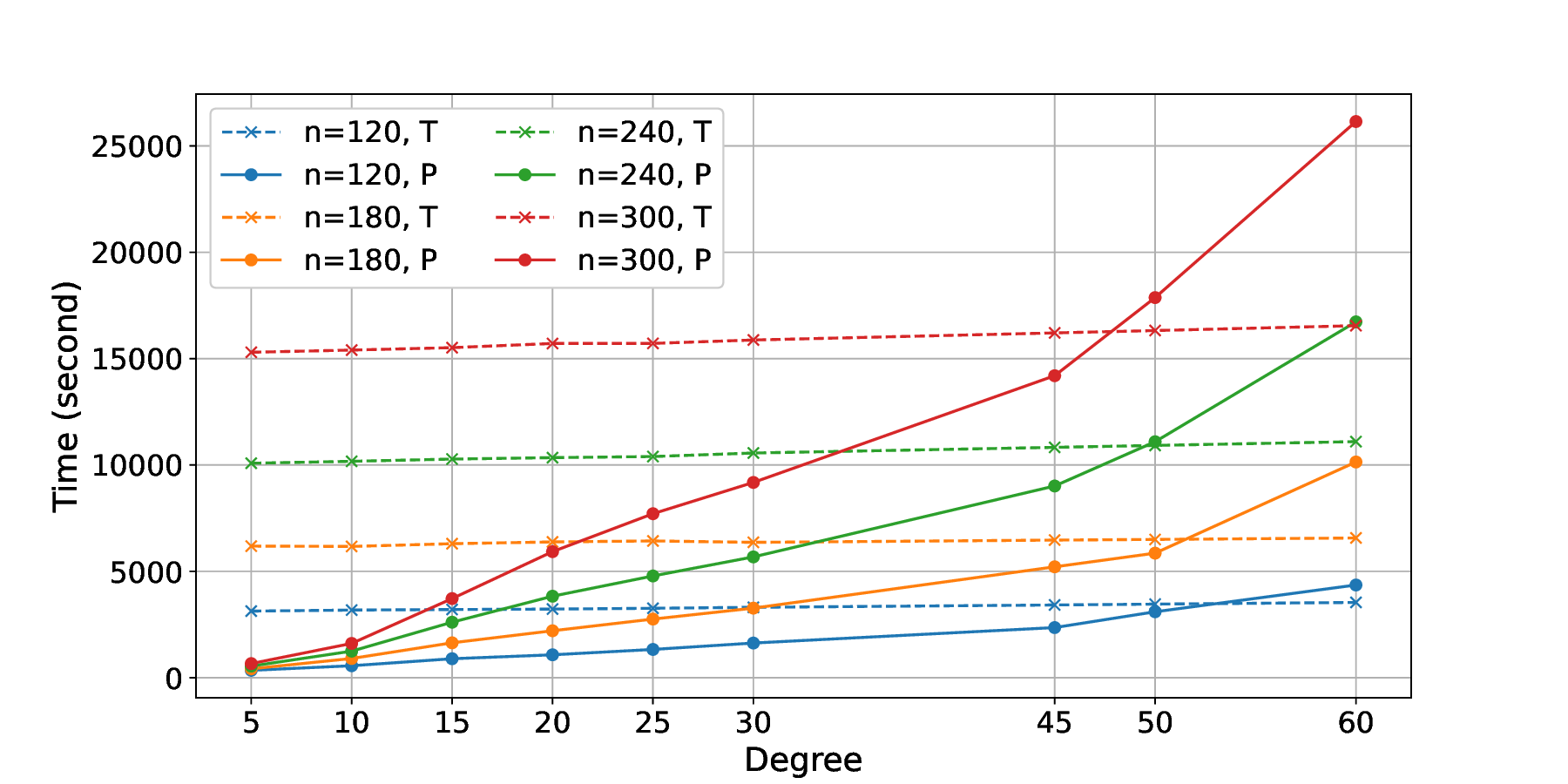}
	\caption{Wall-clock time allocation on Random Regular Graph. P: curvature precomputation phase, T: GCNN training phase.}
	\label{fig:PT_wall-clock}
    \vspace{-0.5cm}
\end{figure}

\begin{figure}
	\centering
	\includegraphics[width=1.0\linewidth]{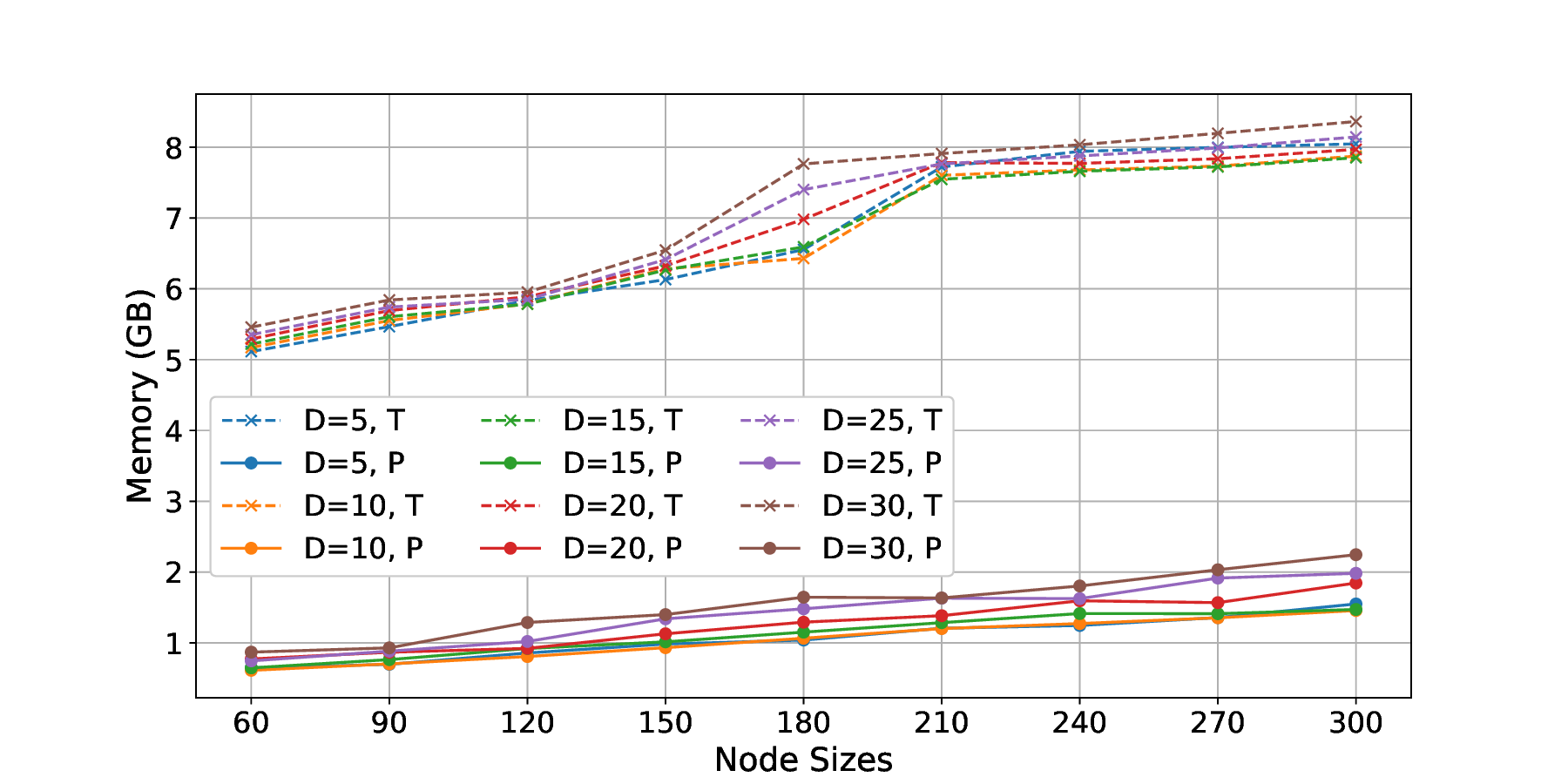}
	\caption{Memory allocation on Random Regular Graph. P: curvature precomputation phase, T: GCNN training phase.}
	\label{fig:PT_memory}
    \vspace{-0.5cm}
\end{figure}

This subsection evaluates the scalability and computational efficiency of RicciPool on COLLAB and REDDIT-M-5K. These two datasets have different graph sizes and densities. COLLAB is highly dense, with an average density of approximately $0.91$, while REDDIT-M-5K contains larger but much sparser graphs, with an average density of approximately $0.0046$. This contrast allows us to examine how graph size and graph density affect the computational overhead of RicciPool, especially the offline curvature precomputation cost.

Table~\ref{tab:time_and_memory} further presents the memory usage and wall-clock time for the curvature precomputation and model training phases on these two large graph datasets. It is noteworthy that on the REDDIT-M-5K dataset, which has a larger number of nodes, the curvature precomputation time is much shorter than the model training time. In contrast, on the relatively denser COLLAB dataset, the curvature precomputation time exceeds the training time. This indicates that curvature precomputation and model training respond differently to graph scale (including the number of nodes and density). To investigate this relationship further, we conducted controlled experiments on random regular graphs. 

We systematically varied the graph scale by adjusting the number of nodes $\{120, 180, 240, 300\}$ and the node degree $\{5, 10, 15, 20, 25, 30, 45, 50, 60\}$. 
For each combination of number of node and node degree, we generated $500$ graphs, and recorded the resource consumption for each during both the curvature precomputation and model training phases. 

The experimental results (Fig.~\ref{fig:PT_wall-clock} and Fig.~\ref{fig:PT_memory}) show that as the number of nodes increases, the time required for both curvature precomputation and model training increases. However, for a fixed number of nodes, as the graph density (node degree) increases, the curvature computation time exhibits an approximate polynomial growth, while the training time increases only linearly. For instance, in a graph with $120$ nodes, when the node degree exceeds $60$ (density $\approx 0.5$), the curvature precomputation time surpasses the model training time. This indicates that in large-scale, high density graph scenarios, the complexity of curvature precomputation may limit the scalability of algorithm.

This scalability limitation is mainly related to graph density. Curvature precomputation becomes more expensive on dense graphs, because more local neighborhood interactions need to be considered during Ollivier Ricci flow. In contrast, for graphs with moderate connectivity, the offline curvature precomputation cost is usually more manageable. In our experiments, the precomputation time is generally lower than the model training time within this range. Therefore, the extra computational cost of RicciPool mainly comes from the offline curvature precomputation stage and is sensitive to graph density.

\begin{table*}[htbp]
\centering
\caption{Wall-clock Time (s) and Memory (GB) Comparison for Training and Precomputation Phase}
\label{tab:time_and_memory}
\setlength{\tabcolsep}{3pt}
\begin{tabular}{lccccccccc}
\toprule
& \multicolumn{3}{c}{\textbf{Training}} & \multicolumn{2}{c}{\textbf{Precomputation (\textit{l}=1)}} & \multicolumn{2}{c}{\textbf{Precomputation (\textit{l}=2)}} & \multicolumn{2}{c}{\textbf{Precomputation (\textit{l}=3)}} \\
\cmidrule(lr){2-4} \cmidrule(lr){5-6} \cmidrule(lr){7-8} \cmidrule(lr){9-10}
\textbf{Dataset} & \textbf{Wall-clock Time} & \textbf{Memory} & \textbf{GPU Memory} & \textbf{Wall-clock Time} & \textbf{Memory} & \textbf{Wall-clock Time} & \textbf{Memory} & \textbf{Wall-clock Time} & \textbf{Memory} \\
\midrule
ENZYMES & 928.95 & 5.1146 & 2.044 & 33.67 & 0.4442 & 69.10 & 0.4387 & 105.68 & 0.4473 \\
D\&D & 30848.21 & 5.4996 & 7.528 & 478.58 & 2.0508 & 944.91 & 2.0306 & 1435.05 & 2.0090 \\
PROTEINS & 1964.74 & 5.1793 & 2.090 & 73.84 & 0.4865 & 145.87 & 0.4973 & 226.66 & 0.4866 \\
IMDB-B & 1416.51 & 5.1338 & 2.012 & 81.03 & 0.4967 & 146.96 & 0.5027 & 217.90 & 0.5023 \\
IMDB-M & 1963.49 & 5.1116 & 2.010 & 107.01 & 0.5074 & 156.90 & 0.5057 & 220.96 & 0.5051 \\
COLLAB & 16988.21 & 9.1408 & 2.554 & 24803.02 & 3.6249 & 25370.60 & 3.6138 & 37287.22 & 3.6086 \\
REDDIT-M-5K &  505102.92 & 7.5728  &  14.086 & 19493.68  & 13.8133 & 39442.66  &  13.9213 &  59093.87 & 13.9824  \\
\bottomrule
\end{tabular}
\end{table*}
\vspace{-0.3cm}
\subsection{Computational Efficiency Analysis}
\label{sec:cost_anslysis}
\begin{figure}
	\centering
	\includegraphics[width=1.0\linewidth]{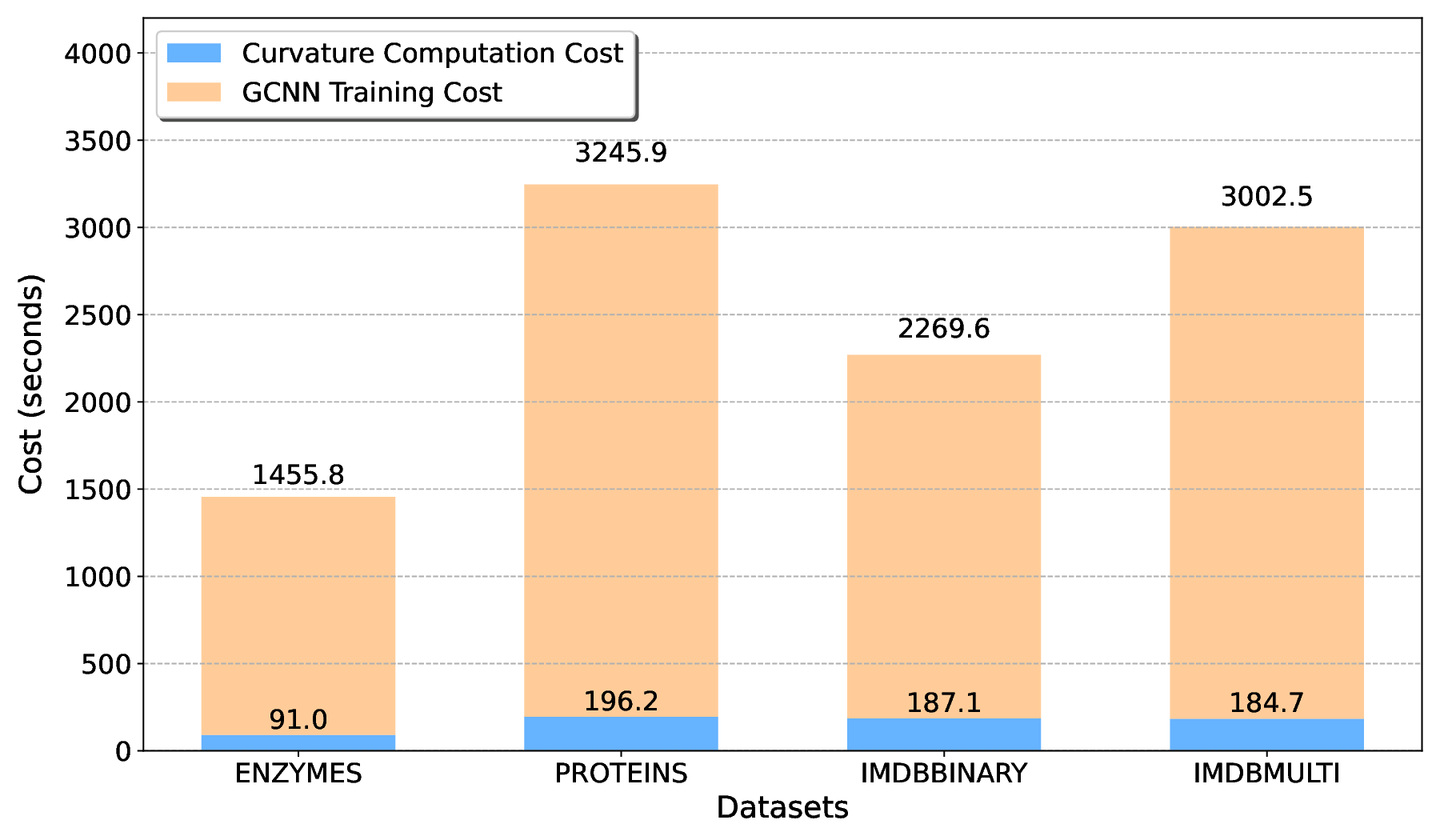}
	\caption{Costs on curvature precomputation and GCNN training.}
	\label{fig:cost}
    \vspace{-0.3cm}
\end{figure}

The computational efficiency challenges inherent in Ollivier-Ricci curvature motivated our design of an optimization mechanism: the offline curvature computation and online pooling decoupling framework. RicciPool adopts a two phase computational paradigm. \textbf{Precomputation Phase}: Discrete Ricci curvature flow iterations are executed to extract curvature features, which is a process entirely independent of the neural network training pipeline. \textbf{Pooling Phase}: Hierarchical pooling is performed based on precomputed curvature features, exhibiting computational cost comparable to standard graph convolutional layers significantly lower than online curvature optimization approaches.

We conducted a comparative analysis between curvature precomputation cost and GCNN training duration. As demonstrated in Fig.~\ref{fig:cost}, for the four datasets considered in this comparison, the curvature precomputation time is lower than the GCNN training time. Taking the PROTEINS dataset as an example, the curvature precomputation time constitutes only $6.25\%$ of the total training time ($91.0$ seconds vs. $1455.8$ seconds). 

Table~\ref{tab:time_and_memory} compares the wall-clock time and peak memory usage between the curvature precomputation phase and the model training phase across the experimental datasets. In terms of memory consumption, the curvature precomputation stage requires less memory than the training stage on all datasets, and this memory usage is independent of the number of Ricci flow iterations. Regarding time cost, the training time significantly exceeds the curvature precomputation time on most datasets. COLLAB is an exception, mainly because its high graph density makes curvature computation over local neighborhoods much more expensive. This result is consistent with the scalability analysis, showing that graph density has a stronger impact on curvature precomputation than on GCNN training.

Overall, the extra computational cost of RicciPool mainly comes from the offline curvature precomputation stage, which is sensitive to graph density. The GCNN training and inference processes remain unchanged, since curvature information is computed before model training and then used to guide graph pooling.

\vspace{-0.3cm}
\subsection{Geometric Interpretability of RicciPool}
We systematically elucidate the mechanistic interpretability of RicciPool through two dimensions: simulation validation and case studies. Simulation experiments explain the relationship between graph curvature and clustering, while real-world case studies demonstrate curvature's impact on clustering outcomes.
\begin{figure}[h]
	\centering
	\includegraphics[width=1.0\linewidth]{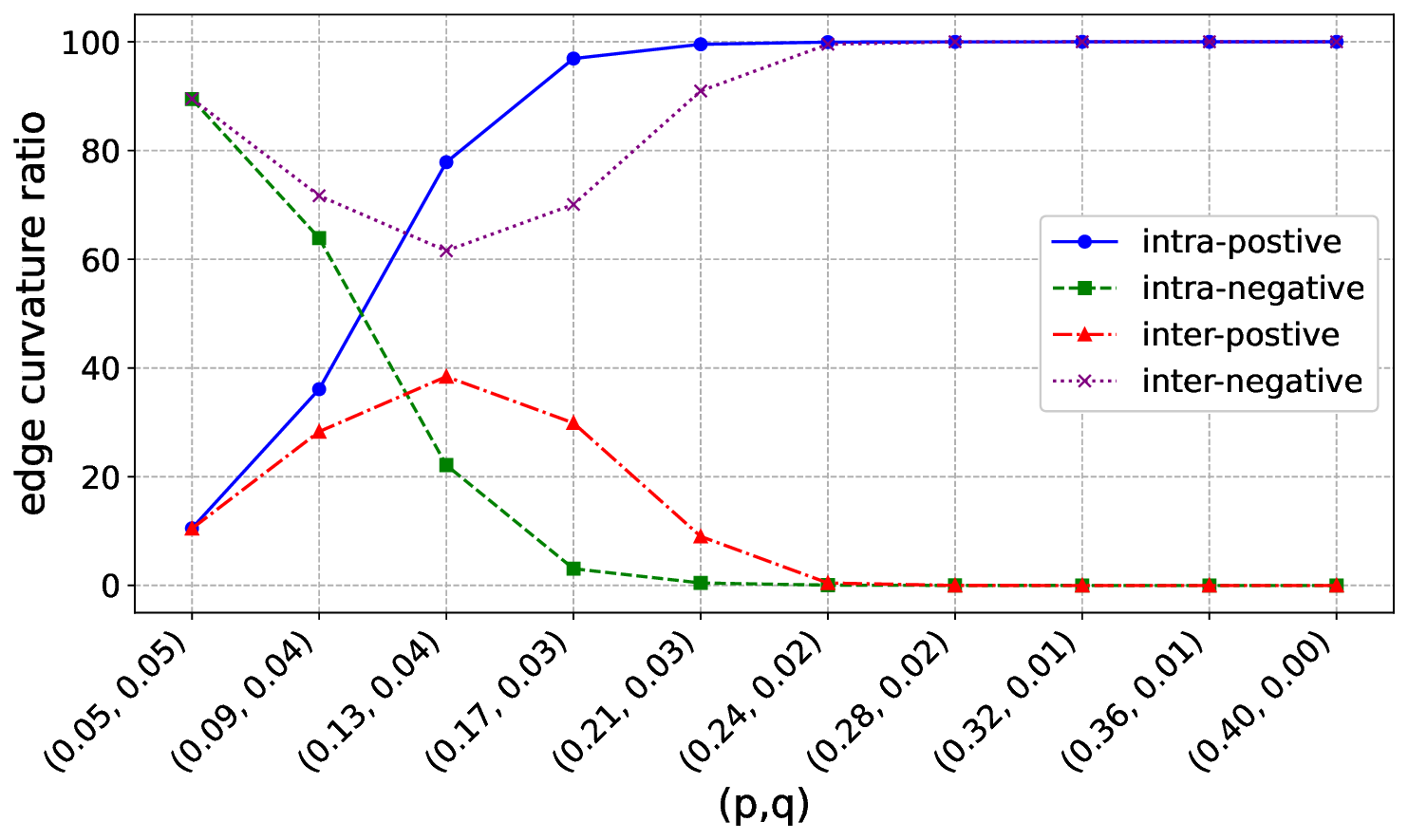}
	\caption{Ollivier-Ricci curvature distribution in different clustering phases. intra-positive: proportion of intra-cluster edges with positive curvature, intra-negative: proportion of intra-cluster edges with negative curvature, inter-positive: proportion of inter-cluster edges with positive curvature, inter-negative: proportion of inter-cluster edges with negative curvature.}
	\label{fig:curvature_cluster}
    \vspace{-0.5cm}
\end{figure}

\textbf{Simulation Experiments}. We simulate the clustering process on a Stochastic Block Model (SBM) to observe changes in Ollivier-Ricci curvature distribution under different clustering states.
SBM model has two key parameters: $p$ is the intra-cluster connection probability to control cluster tightness and $q$ is inter-cluster connection probability (controls separation between clusters). Therefore, we can simulate this process by adjusting the values of $p$ and $q$. 

Clustering Simulation: The clustering process is simulated by gradually increasing $p$ (from $0.05$ loose to $0.40$ tight) while decreasing $q$ (from $0.05$ to $0.0$). Initial $q = p$ implies indistinguishable intra/inter-cluster connections at the initial clustering stage.

Curvature Distribution Findings (shown in Fig.~\ref{fig:curvature_cluster}): for intra-cluster edges, the proportion of positive-curvature edges increases as clustering progresses, the proportion of negative-curvature edges decreases, and finally converges to $100\%$ positive curvature. For inter-cluster edges, the proportion of positive-curvature edges increases initially, then decreases; the proportion of negative-curvature edges decreases initially, then increases, and finally converges to $100\%$ negative curvature. 
Therefore, our simulation suggests a clear relationship between the sign of curvature and cluster cohesion: positive curvature becomes predominant within clusters, while negative curvature becomes predominant between clusters as separation increases. The prevalence of negative-curvature edges between clusters in well-separated SBM graphs is consistent with the role of Ricci flow in refining cluster boundaries, as observed in our graph pooling experiments.

\textbf{Case Study}. We analyze real graphs to observe curvature-driven changes in edge weights, shown in Fig.~\ref{fig:r1}. Edge weights (red) and curvature values (blue) are annotated. We can clearly see that positive curvature can promote tighter connections for intra-cluster edges (taking edge $\langle 6,7 \rangle$  in the ENZYMES graph $351$ as an example, the distances after curvature flow iterations $1$ and $2$ are $0.5$ and $0.21$, respectively). Negative curvature can promote looser connections for inter-cluster edges (taking edge $\langle 1,2 \rangle$ in the PROTEINS graph $1084$ as an example, the distances after curvature flow iterations $1$ and $2$ are $1.33$ and $1.85$, respectively).

\subsection{Algorithm Stability and Sensitivity Analysis}

\begin{figure}
	\centering
	\includegraphics[width=1.0\linewidth]{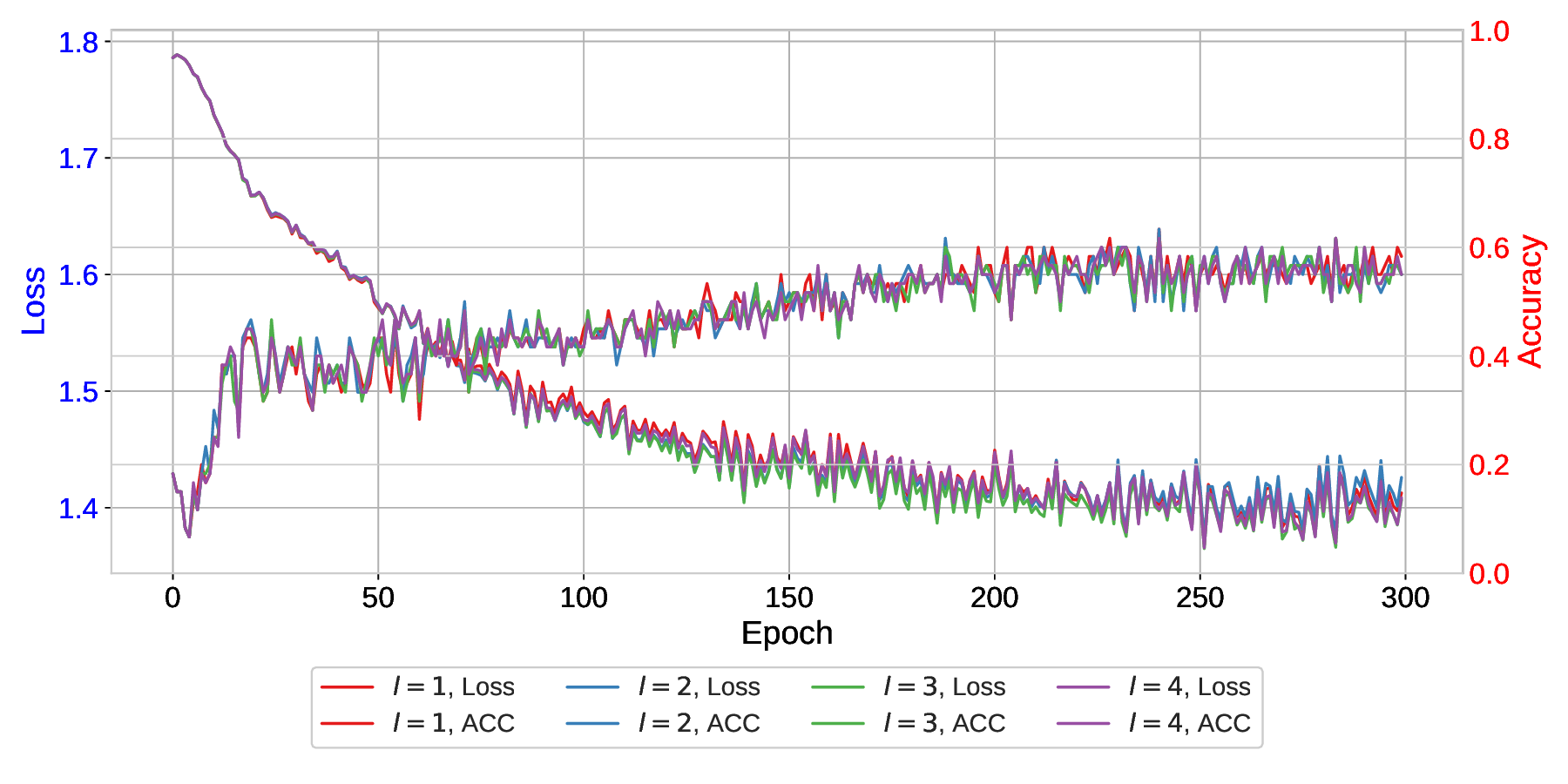}
	\caption{Trends in validation loss and accuracy in different ricci flow iteration $l$ on ENZYMES dataset.}
	\label{fig:ENZYMES_loss_acc_ricci_iters}
    \vspace{-0.5cm}
\end{figure}

\begin{figure}
	\centering
	\includegraphics[width=1.0\linewidth]{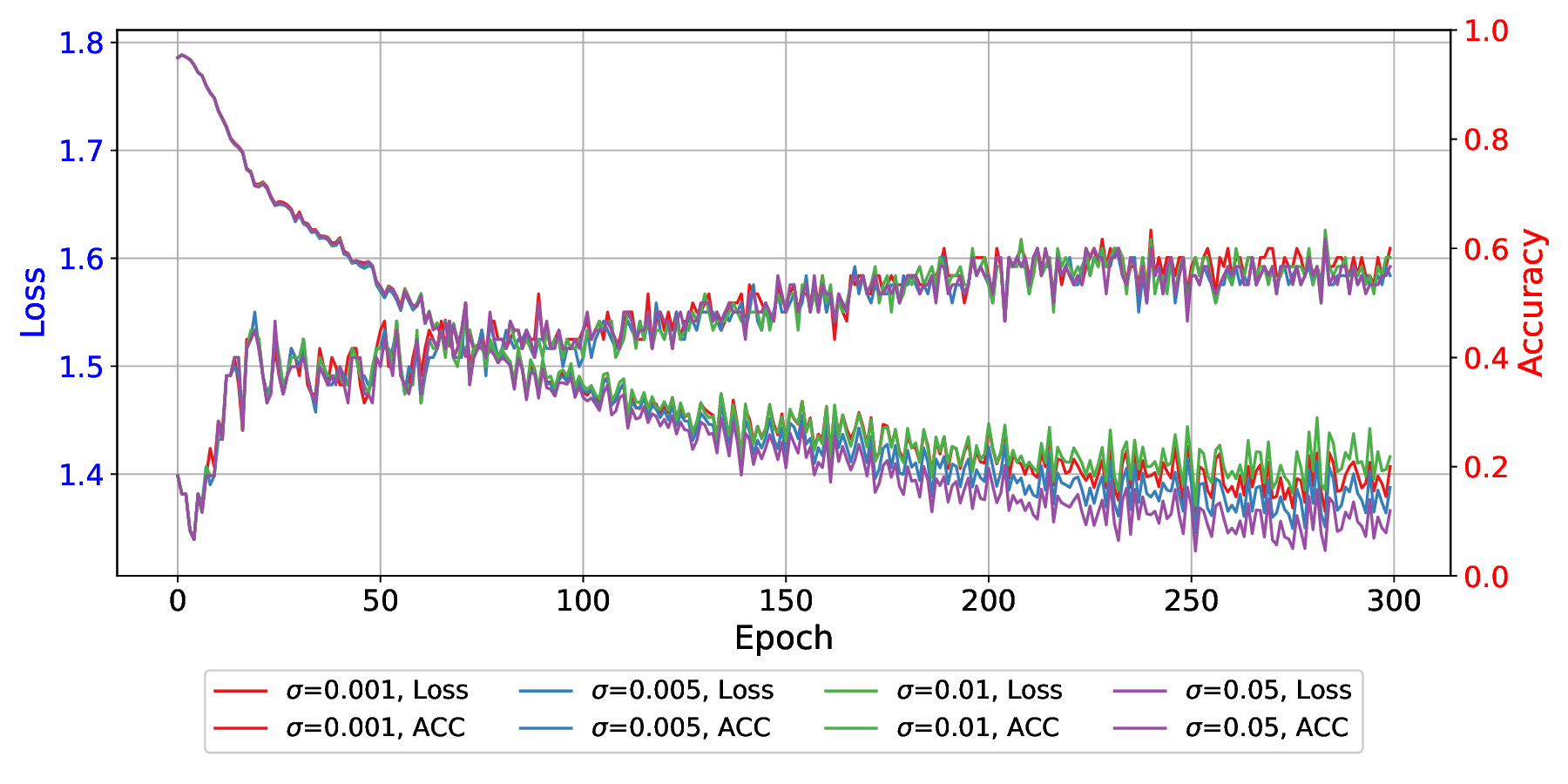}
	\caption{Trends in validation loss and accuracy in different stop threshold $\delta$ on ENZYMES dataset.}
	\label{fig:ENZYMES_loss_acc_ricci_stop_thre}
    \vspace{-0.5cm}
\end{figure}

To evaluate the stability and sensitivity of the curvature flow-based graph pooling method, experiments were conducted on the ENZYMES and IMDB-B datasets. The maximum number of training epochs was set to $300$ for ENZYMES and $500$ for IMDB-B (show in Appendix~\ref{appendix:Sensitivity_IMDBB}), while the number of curvature flow iterations was varied as $\{1, 2, 3, 4\}$. Fig.~\ref{fig:ENZYMES_loss_acc_ricci_iters} illustrates the evolution of validation loss and accuracy during training. As observed, the loss curves decrease steadily under different iteration numbers, indicating that RicciPool maintains stable optimization when the iteration number changes within a moderate range. 

We further examine the effect of the stopping threshold. The threshold is defined by the mean curvature change and is varied in \(\{0.001, 0.005, 0.01, 0.05\}\). The results, presented in Fig.~\ref{fig:ENZYMES_loss_acc_ricci_stop_thre}, indicate that accuracy remains largely stable, exhibiting only minor fluctuations and forming a clear plateau across a broad intermediate range of threshold values. One possible reason is that RicciPool does not require exact curvature convergence. Once the curvature flow has provided sufficient structural separation for node clustering, small changes in the stopping threshold have limited influence on the relative edge weight patterns and the learned cluster assignment matrix. Therefore, a threshold selected from this stable range can provide reliable performance while avoiding unnecessary curvature iterations.
\section{Conclusion}\label{sec:conc}

Graph pooling operations are essential for managing large-scale graph-structured datasets. Recently, several advanced methods have been proposed, yet they often overlook higher-order spatial connections among neighboring nodes, focusing primarily on rough graph topology. Addressing these limitations, we introduce a novel graph pooling approach using discrete graph Ricci flow. Our method treats graph pooling as a node clustering task, leveraging Ollivier-Ricci curvature to distinguish well-connected subgraphs from tree-like structures. We employ spectral clustering theory to learn a cluster assignment matrix, ensuring that the coarse graph after pooling preserves the original global topology. 
Experiments on benchmark datasets demonstrate the effectiveness and competitiveness of RicciPool.

Furthermore, our method sheds light on specific phenomena. For instance, in fully connected graphs where all edge weights and Ollivier-Ricci curvatures are uniform, RicciPool cannot uncover a more compact latent structure, limiting its ability to enhance graph representations for high-level tasks. While clustering nodes in a fully connected graph may seem impractical due to uniform importance scores, our method effectively uses Ollivier-Ricci curvature to identify these structures. Thus, our approach provides robust geometric interpretability, offering theoretical insights into these observations.

In the future, we will continue to explore the application potential of geometric flow within GNNs. For instance, Ricci flow can automatically identify the clustering structure of graph node sets. A key question then arises: how can geometric flow be leveraged to autonomously determine the optimal number of clusters? Furthermore, the evolution of geometric flows on graph data exhibits parallels with GNN dynamics: node clustering resembles the GNN over-smoothing phenomenon, while the inter-class separation driven by geometric flow evolution mirrors the GNN over-squashing process. Consequently, our subsequent work will focus on an in-depth investigation of the relationship between the Ricci flow evolution process and these two specific GNN phenomena, aiming to establish a more robust theoretical explanation.

\appendix

\subsection{Stability and Sensitivity for IMDB-B}
\begin{figure}
	\centering
	\includegraphics[width=1.0\linewidth]{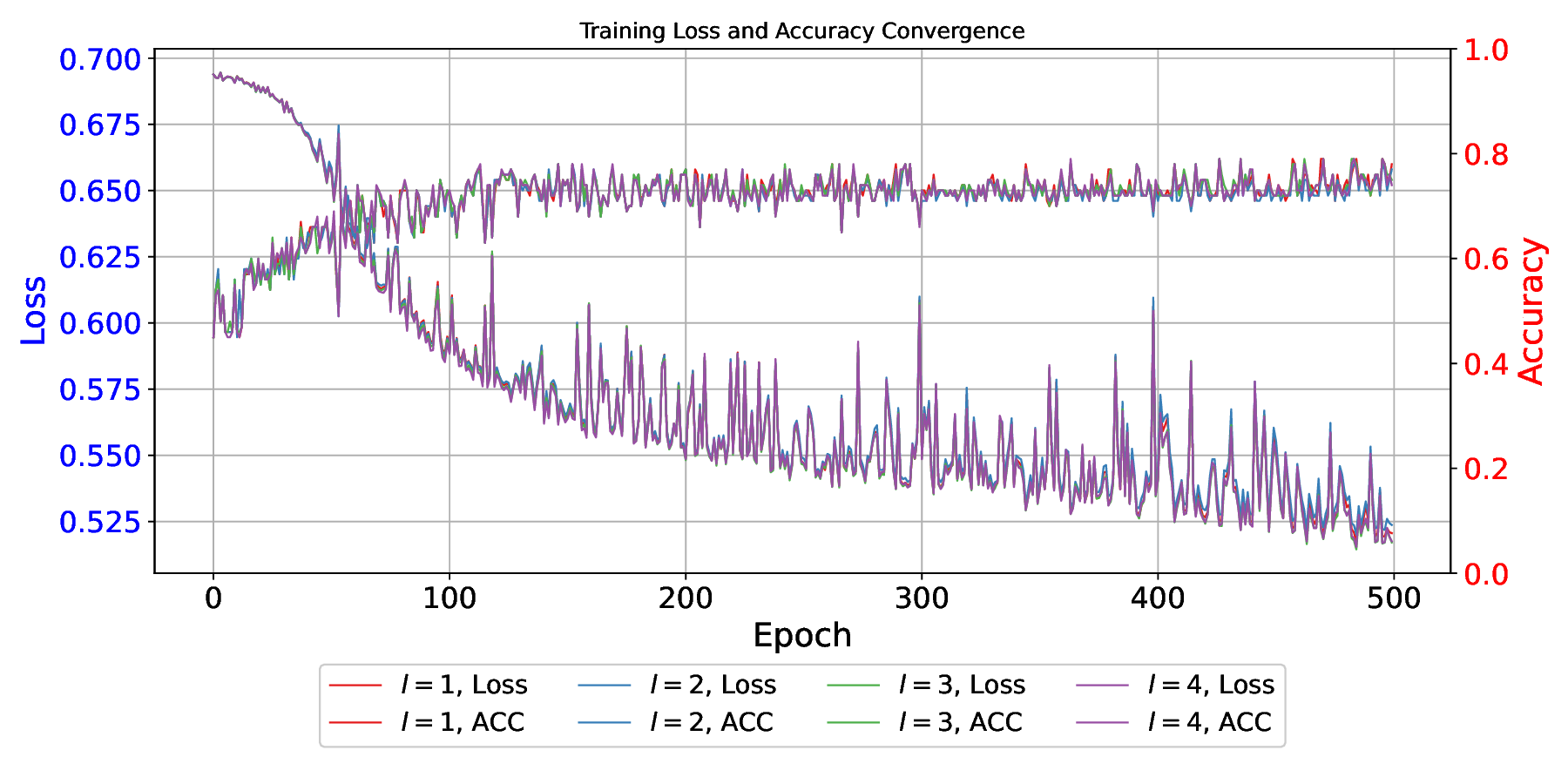}
	\caption{Trends in validation loss and accuracy in different ricci flow iteration $l$ on IMDB-B dataset.}
	\label{fig:Sensitivity_IMDBB}
\end{figure}

\begin{figure}
	\centering
	\includegraphics[width=1.0\linewidth]{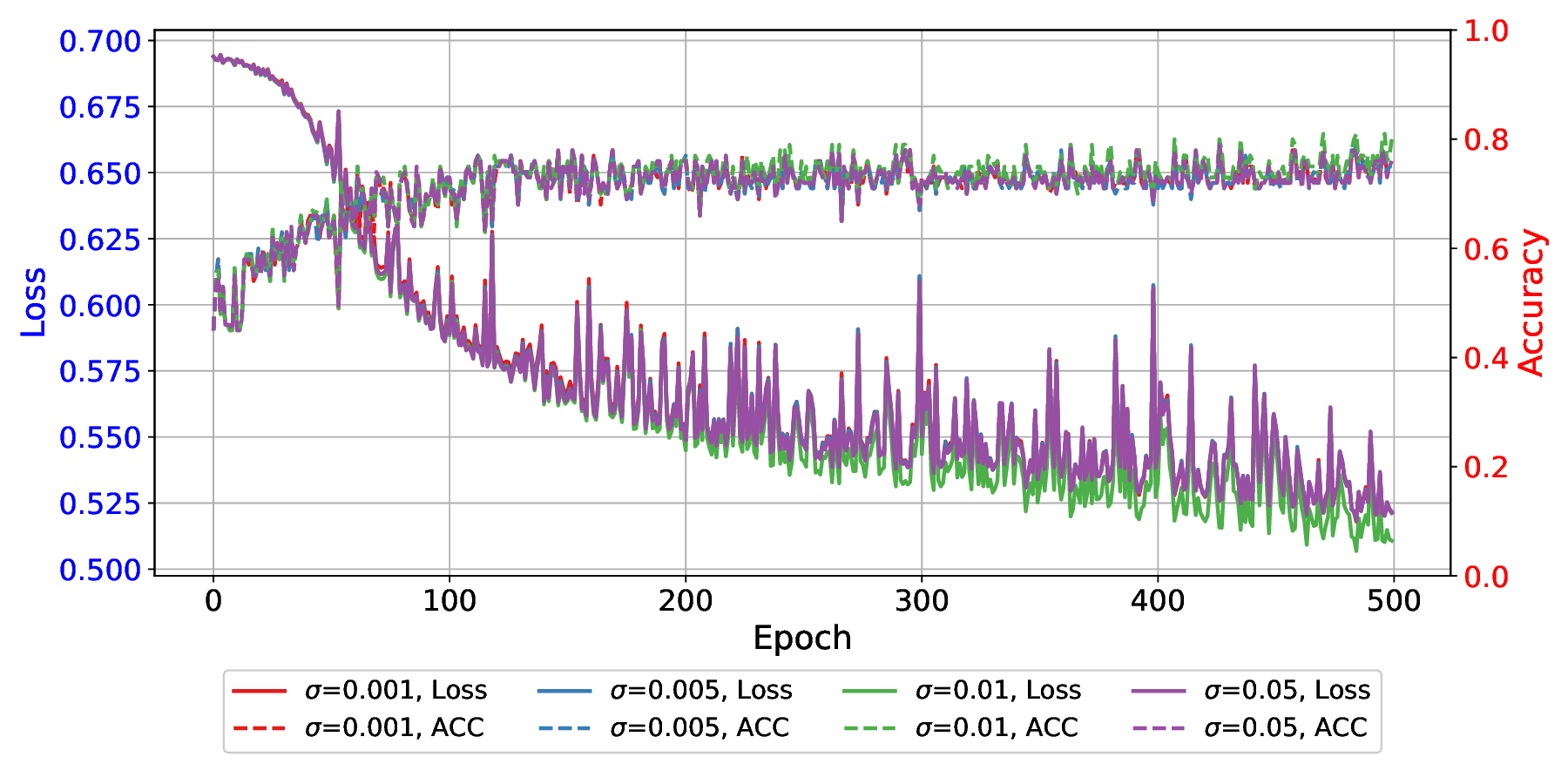}
	\caption{Trends in validation loss and accuracy in different stop threshold $\delta$ on ENZYMES dataset.}
	\label{fig:Stop_thre_IMDBB}
\end{figure}

Fig.~\ref{fig:Sensitivity_IMDBB} and Fig.~\ref{fig:Stop_thre_IMDBB} present the experimental results on the stability and sensitivity of RicciPool on the IMDB-B dataset. The results indicate that, relative to the ENZYMES dataset, IMDB-B exhibits greater stability in its loss and accuracy convergence curves across different numbers of curvature flow iterations. Furthermore, its training trajectory shows marked insensitivity to variations in the stopping threshold.

\label{appendix:Sensitivity_IMDBB}

\begin{thebibliography}{00}

\bibitem{spectralpool}
Bianchi, F. M., Grattarola, D., and Alippi, C.. Spectral clustering with graph neural networks for graph pooling. In {\it International Conference on Machine Learning}, 2020.
\bibitem{Anca2009}
Bonciocat, A. I., and Sturm, K. T.. Mass transportation and rough curvature bounds for discrete spaces. {\it Journal of Functional Analysis}, 256(9):2944–2966, 2009.
\bibitem{Karsten2005}
Borgwardt, K.M., Ong, C. S., Schonauer, S., Vishwanathan, S. V. N., Smola, A. J., and Kriegel, H. P.. Protein function prediction via graph kernels. {\it Bioinformatics,} 21(suppl-1): i47-i56, 2005.
\bibitem{bronstein2017geometric}
Bronstein, M. M., Bruna, J., LeCun, Y., Szlam, A., and Vandergheynst, P.. Geometric deep learning: going beyond euclidean data. {\it IEEE Signal Processing Magazine}, 34(4):18–42, 2017.
\bibitem{Joan2013}
Bruna, J., Zaremba, W., Szlam, A., and LeCun, Y.. Spectral networks and locally
connected networks on graphs. {\it arXiv preprint arXiv:1312.6203}, 2013.
\bibitem{cai2018}
Cai, H., Zheng, V. W., and Chang, K. C. C.. A comprehensive survey of graph embedding: problems, techniques, and applications. {\it IEEE Transactions on Knowledge and Data Engineering}, 30, 9, pp. 1616-1637, 2018.
\bibitem{Wu_and_Pan_2021}
Z.Wu, S. Pan, F. Chen, G. Long, C. Zhang and P. S. Yu. A Comprehensive Survey on Graph Neural Networks. In {\it IEEE Transactions on Neural Networks and Learning Systems}, 32(1):4-24, 2021. 
\bibitem{cai2020}
Cai, L., and Ji, S.. A multi-scale approach for graph link prediction. In {\it Thirty-Fourth AAAI Conference on Artificial Intelligence}, 2020.
\bibitem{defferrard2016}
Defferrard, M., Bresson, X., and Vandergheynst, P.. Convolutional neural networks on graphs with fast localized spectral filtering. In {\it Advances in Neural Information Processing Systems}, pp. 3844–3852, 2016.
\bibitem{D&D2003}
Dobson, P. D., and Doig, A. J.. Distinguishing enzyme structures from non-enzymes without alignments {\it Journal of molecular biology}, 330(4):771-783, 2003.
\bibitem{Forman2003} 
Forman, R.. Bochner’s method for cell complexes and combinatorial ricci curvature. {\it Discrete and Computational Geometry}, 29(3):323–374, 2003.
\bibitem{gao2019b}
Gao, H., and Ji, S.. Graph representation learning via hard and channel-wise attention networks. In {\it Proceedings of the 25th ACM SIGKDD International Conference on Knowledge Discovery \& Data Mining}, pp. 741–749, 2019.
\bibitem{gao2019} 
Gao, H., and Ji, S.. Graph u-nets. In {\it International Conference on Machine Learning}, pp. 2083–2092, 2019.

\bibitem{will2017}
Hamilton, W., Ying, Z., and Leskovec, J.. Inductive representation learning on  large graphs. In {\it Advances in Neural Information Processing Systems}, 2017.
\bibitem{hamilton1982three} 
Hamilton, R. S. Three-manifolds with positive ricci curvature. {\it Journal of Differential Geometry}, 17(2):255–306, 1982.
\bibitem{Henaff2015}
Henaff, M., Bruna, J., and LeCun, Y.. Deep convolutional networks on graph-structured data. {\it arXiv preprint arXiv:1506.05163}, 2015.
\bibitem{selfpool} 
Lee, J., Lee, I., and Kang, J.. Self-attention graph pooling. In {\it Proceedings of the 36th International Conference on Machine learning (ICML-19)}, 2019.
\bibitem{Grattarola_and_Zambon}
D. Grattarola, D. Zambon, F. M. Bianchi and C. Alippi. Understanding Pooling in Graph Neural Networks. In {\it IEEE Transactions on Neural Networks and Learning Systems}, 35(2):2708-2718, 2024.
\bibitem{Wang_and_Chang_2022}
Y.Wang, D. Chang, Z. Fu and Y. Zhao. Seeing All From a Few: Nodes Selection Using Graph Pooling for Graph Clustering. In {\it IEEE Transactions on Neural Networks and Learning Systems}, 35(5):7231-7237, 2024. 
\bibitem{Bianchi_and_Grattarola_2022}
F. M. Bianchi, D. Grattarola, L. Livi and C. Alippi. Hierarchical Representation Learning in Graph Neural Networks With Node Decimation Pooling. In {\it IEEE Transactions on Neural Networks and Learning Systems}, 33(5):2195-2207, 2022. 
\bibitem{li2022curvature} 
Li, H., Cao, J., Zhu, J., Liu, Y., Zhu, Q., and Wu, G.. Curvature graph neural network. {\it Information Sciences}, 592:50–66, 2022.
\bibitem{Lin2011} 
Lin, Y., Lu, L., and Yau, S. T.. Ricci curvature of graphs. {\it Tohoku Mathematical Journal, Second Series}, 63(4):605–627, 2011.
\bibitem{eigenpool} 
Ma, Y., Wang, S., Aggarwal, C. C., and Tang, J.. Graph convolutional networks with eigenpooling. In {\it Proceedings of the 25th ACM SIGKDD International Conference on Knowledge Discovery \& Data Mining}, pp:723-731, 2019.
\bibitem{Monti2017}
Monti, F., Boscaini, D., Masci, J., Rodola, E., Svoboda, J., and Bronstein, M. M.. Geometric deep learning on graphs and manifolds using mixture model cnns. In {\it Proceedings of the IEEE Conference on Computer Vision and Pattern Recognition}, pp. 5115–5124, 2017.
\bibitem{ni2018network}
Ni, C.-C., Lin, Y.-Y., Gao, J., and Gu, X.. Network alignment by discrete ollivier-ricci flow. In {\it International Symposium on Graph Drawing and Network Visualization}, pp. 447–462. Springer, 2018.
\bibitem{ni2019community} 
Ni, C.-C., Lin, Y.-Y., Luo, F., and Gao, J.. Community detection on networks with ricci flow. {\it Scientific Reports}, 9(1):1–12, 2019.
\bibitem{ollivier2009ricci} 
Ollivier, Y.. Ricci curvature of markov chains on metric
spaces. {\it Journal of Functional Analysis}, 256(3):810–864, 2009.
\bibitem{ollivier2010}
Ollivier, Y.. A survey of ricci curvature for metric spaces and markov chains. {\it In Probabilistic Approach to Geometry}, 343-381, 2010.
\bibitem{samal2018}
Samal, A., Sreejith, R. P., Gu, J., Liu, S., Saucan, E., and Jost, J.. Comparative analysis of two discretizations of Ricci curvature for complex networks. {\it Sci. Rep. 8}, 8650, 2018.
\bibitem{sandhu2015}
Sandhu, R., Georgiou, T., Reznik, E., Zhu, L., Kolesov, I., Senbabaoglu, Y., and Tannenbaum, A.. Graph curvature for differentiating cancer networks. {\it Sci. Rep. 5}, 12323, 2015.
\bibitem{Yu2003}
Shi. Multiclass spectral clustering. In {\it Proceedings Ninth IEEE International Conference on Computer Vision}, pp.313-319 vol.1, Oct 2003.
\bibitem{sia2019}
Sia, J., Jonckheere, E., and Bogdan, P.. Ollivier-ricci curvature-based method to community detection in complex networks. {\it Scientific Reports}, 9(1):1–12, 2019.
\bibitem{siddharth2021}
Siddharth P., Feng Y., Terrence J., Ram R., Amotz B. and Ananthram S.. An efficient alternative to Ollivier-Ricci curvature based on the Jaccard metric. {\it arXiv:1710.01724v1}, 2021.
\bibitem{sreejith2016}
Sreejith, R. P., Mohanraj, K., Jost, J., Saucan, E. and Samal, A.. Forman curvature for complex networks. {\it J. Stat. Mech: Theory Exp. 2016}, 063206, 2016.
\bibitem{Petar2018} 
Velickovic, P., Cucurull, G., Casanova, A., Romero, A., Lio, P., and Bengio, Y.. Graph attention networks. In {\it International Conference on Learning Representations}, 2018.
\bibitem{haarpool}
Wang, Y. G., Li, M., Ma, Z., Montufar, G., Zhuang, X., and Fan, Y.. Haar graph pooling.  In {\it International Conference on Machine Learning}, 2020.
\bibitem{weber2018} 
Weber, M., Jost, J., and Saucan, E.. Detecting the coarse geometry of networks. In {\it NeurIPS 2018 Workshop}, 2018.
\bibitem{weber2016} 
Weber, M., Jost, J., and Saucan, E.. Forman-ricci flow for change detection in large dynamic data sets. {\it Axioms}, 5(4):26, 2016.
\bibitem{weber2017} 
Weber, M., Saucan, E., and Jost, J.. Characterizing complex networks with forman-ricci curvature and associated geometric flows. {\it Journal of Complex Networks}, 5(4):527–550, 2017.
\bibitem{SEP} Wu, J., Chen, X., Xu, K., and Li, S.. Structural entropy guided graph hierarchical pooling.In {\it International conference on machine learning}. ICML, 2022: 24017-24030.
\bibitem{Xu2019}
Xu, K., Hu, W., Leskovec, J., and Jegelka, S.. How powerful are graph neural
networks? In {\it International Conference on Learning Representations}, 2019.
\bibitem{Pinar2015a} 
Yanardag, P., and Vishwanathan, S. V. N.. A structural smoothing framework for robust graph comparison. In {\it Advances in Neural Information Processing Systems}, pp.2134-2142, 2015a.
\bibitem{Pinar2015b}
Yanardag, P., and Vishwanathan, S. V. N.. Deep graph kernels. In {\it Proceedings of the 21th ACM SIGKDD International Conference on Knowledge Discovery and Data Mining}, pp.1365-1374, ACM, 2015b.
\bibitem{ye2019curvature}
Ye, Z., Liu, K. S., Ma, T., Gao, J., and Chen, C. Curvature
graph network. In {\it International Conference on Learning Representations}, 2019.
\bibitem{diffpool}
Ying, Z., You, J., Morris, C., Ren, X., Hamilton, W., and Leskovec, J.. Hierarchical graph representation learning with differentiable pooling. In {\it Advances in Neural Information Processing Systems}, pp. 4800–4810, 2018.
\bibitem{structpool}
Yuan, H., and Ji, S.. Structpool: Structured graph pooling via conditional random fields. In {\it Proceedings of the 8th International Conference on Learning Representations}, 2020.
\bibitem{zhang2018linkprediction}
Zhang, M., and Chen, Y.. Link prediction based on graph neural networks. In {\it Advances in Neural Information Processing Systems}, pp. 5165–5175, 2018.
\bibitem{zhang2018graphclassification} 
Zhang, M., Cui, Z., Neumann, M., and Chen, Y.. An end-to-end deep learning architecture for graph classification. In {\it AAAI}, pp. 4438–4445, 2018.
\bibitem{orc_arxiv}
Amy F and Melanie W. Graph Pooling via Ricci Flow. {\it Transactions on Machine Learning Research}, 2024.
\bibitem{collab}
P. Yanardag and S. V. N. Vishwanathan. A structural smoothing framework for robust graph
comparison. In {\it Advances in Neural Information Processing Systems}, 28:2134–2142, 2015.
\bibitem{reddit-multi-5k}
Yanardag, Pinar and Vishwanathan, S.V.N. Deep Graph Kernels. 
In {\it Proceedings of the 21th ACM SIGKDD International Conference on Knowledge Discovery and Data Mining}, 10:1365–1374, 2015. 
\bibitem{CCP-GNN}
P. Zhu, J. Li, Z. Dong, Q. Hu, X. Wang and Q. Wang. CCP-GNN: Competitive Covariance Pooling for Improving Graph Neural Networks. In {\it IEEE Transactions on Neural Networks and Learning Systems}, 36(4):6395-6406, 2025.
\bibitem{MID}
C. Liu, Y. Zhan, B. Yu, L. Liu, B. Du, W. Hu, T. Liu. On exploring node-feature and graph-structure diversities for node drop graph pooling. In {\it Neural Networks}, 167:559-571, 2023.
\bibitem{GAEP}
K. Limbeck, L. Mezrag, G. Wolf, B. Rieck. Geometry-Aware Edge Pooling for Graph Neural Networks. {\it Advances in Neural Information Processing Systems}, 2025. 
\end{thebibliography}
\end{document}